\documentclass[review=false,onefignum,onetabnum]{siamart171218}

\usepackage{lipsum}
\usepackage{amsfonts}
\usepackage{graphicx}
\usepackage{epstopdf}
\usepackage{algorithmic}
\ifpdf
  \DeclareGraphicsExtensions{.eps,.pdf,.png,.jpg}
\else
  \DeclareGraphicsExtensions{.eps}
\fi

\newsiamremark{remark}{Remark}
\newsiamremark{hypothesis}{Hypothesis}
\crefname{hypothesis}{Hypothesis}{Hypotheses}
\newsiamthm{claim}{Claim}

\headers{ML of Space-Fractional PDEs}{M. Gulian, M. Raissi, P. Perdikaris, G. Karniadakis}

\title{Machine Learning of Space-Fractional Differential Equations\thanks{
Submitted to the editors August 2, 2018.
\funding{This work was funded by DARPA EQUiPS grant N66001-15-2-4055, the MURI/ARO grant W911NF-15-1-0562, the AFOSR grant FA9550-17-1-001, and the NSF Graduate Research Fellowship Program.}}}

\author{Mamikon Gulian\thanks{Department of Applied Mathematics, Brown University.}
\and Maziar Raissi\footnotemark[2]
\and Paris Perdikaris\thanks{Department of Mechanical Engineering and Applied Mechanics, School of Engineering and Applied Science, University of Pennsylvania.}
\and George Karniadakis\footnotemark[2]}

\usepackage{amsopn}

 \usepackage{multirow}
\usepackage{bm}
\usepackage{bbm}
\usepackage{amsmath,amssymb}
\usepackage{mathtools}
\usepackage{placeins}
\usepackage{subcaption}
\usepackage{array}
\ifpdf
\hypersetup{
  pdftitle={Machine Learning of Space-Fractional Differential Equations},
  pdfauthor={Mamikon Gulian, Maziar Raissi, Paris Perdikaris, George Karniadakis}
}
\fi

\begin{document}

\newcommand*{\defeq}{\stackrel{\text{def}}{=}}

\maketitle
\renewcommand\vec{\bm}

% REQUIRED
\begin{abstract}
Data-driven discovery of ``hidden physics'' -- i.e., machine learning of differential equation models underlying observed data -- has recently been approached by embedding the discovery problem into a Gaussian Process regression of spatial data, treating and discovering unknown equation parameters as hyperparameters of a ``physics informed'' Gaussian Process kernel. This kernel includes the parametrized differential operators applied to a prior covariance kernel. 
We extend this framework to the  data-driven discovery  of linear \emph{space-fractional} differential equations. The methodology is compatible with a wide variety of space-fractional operators in $\mathbb{R}^d$ and stationary covariance kernels, including the Mat\'ern class, and allows for optimizing the Mat\'ern parameter during training. {Since fractional derivatives are typically not given by closed-form analytic expressions, the main challenges to be addressed are a user-friendly, general way to set up fractional-order derivatives of covariance kernels, together with feasible and robust numerical methods for such implementations.} Making use of the simple Fourier-space representation of space-fractional derivatives in $\mathbb{R}^d$, we provide a unified set of integral formulas for the resulting Gaussian Process kernels. The shift property of the Fourier transform results in formulas involving $d$-dimensional integrals that can be efficiently treated using generalized Gauss-Laguerre quadrature.

The implementation of fractional derivatives has several benefits. First, the method allows for discovering models involving fractional-order PDEs for systems characterized by heavy tails or anomalous diffusion, while bypassing the analytical difficulty of fractional calculus. Data sets exhibiting such features are of increasing prevalence in physical and financial domains. Second,  a single fractional-order archetype allows for a derivative term of arbitrary order to be learned, with the order itself being a parameter in the regression. As a result, even when used for discovering integer-order equations, the proposed method has several benefits compared to previous works on data-driven discovery of differential equations; the user is not required to assume a ``dictionary'' of derivatives of various orders, and directly controls the parsimony of the models being discovered. We illustrate our method on several examples, including fractional-order interpolation of advection-diffusion and modeling relative stock performance in the S\&P 500 with $\alpha$-stable motion via a fractional diffusion equation. 
\end{abstract}

% REQUIRED

% REQUIRED

\section{Introduction}
A novel use of machine learning, which has potential both for modeling with large or high-frequency data sets as well as advancing fundamental science, is the discovery of governing differential equations from data. Unlike highly specialized algorithms used to refine existing models, these novel methods are distinguished by comparatively limited assumptions, and the ability to produce various types of equations from a wide variety of data.

We now give a very brief (and incomplete) overview of a few proposed algorithms. Key differences between various works include the techniques used to generate candidate equations and the technique used to select equations. A fundamental problem that all these work address is that, if such algorithms are to mimic a human scientist, while generating accurate models, they must also avoid or penalize spurious ``overfitted'' models.
In \cite{Schmidt81}, a symbolic regression method was developed to learn conservation laws of physical systems. The laws, which could be nonlinear, were created using genetic programming and evaluated concurrently on predicative ability and parsimony (number of terms) in order to prevent overfitting. 
Earlier, in \cite{Bongard9943} in a dynamical system context, the equations were evaluated using ``probing'' tests while the overfitting was addressed using a separate ``snipping'' process. 
In \cite{Brunton3932}, a more parametric method for learning nonlinear autonomous systems was developed in which the candidate equation is build from a linear combination of elements from a user-defined ``dictionary''. Parsimony translated to sparsity in the dictionary elements, so sparse regression was used to determine the coefficients. 
In \cite{Rudye1602614}, a similar approach was used to generate nonlinear partial differential evolution equations. 

Building on an earlier work \cite{RAISSI2017736} where Gaussian process regression was used to infer solutions of equations, a Gaussian process framework was developed in \cite{RAISSI2017683} for parametric learning of linear differential equations, of the form
\begin{equation} 
\label{linear_relationship}
\mathcal{L}^{p_1,...,p_k} u = f
\end{equation}
given small data on $f$ and $u$. Here $\mathcal{L}^{p_1,...,p_k}$ refers to a linear differential operator with parameters $p_1,...,p_k$; for example, 
$\mathcal{L}^{p_1,...,p_k}$
 could be a linear combination of 
$k$ 
differential operators with coefficients 
$p_1, ..., p_k$. 
The method placed a Gaussian Process on the function $u$ and used linearity of $\mathcal{L}^{p_1,...,p_k}$ to obtain a joint Gaussian Process on $(u,f)$ in which the unknown parameters $p_1,...,p_k$ were subsumed as hyperparameters (this is reviewed in detail in Section \ref{framework}). This allowed the use of standard continuous optimization techniques to optimize the negative log-marginal likelihood, which effects an automatic trade-off between data-fit and model complexity, and thereby find $p_1,..,p_k$. Thus, in this approach, the problem of identifying the differential equation is ``embedded'' into the problem of interpolation/regression of  data. Moreover, the Gaussian Process method only requires computation of the forward action of $\mathcal{L}^{\bm{p}}$ on the covariance kernel, rather than the solution of the differential equation. However, the method also required the user to select a ``dictionary'' of terms and assume a parametric form of the equation.  

The inclusion of fractional-order operators in this framework, with the order itself is a parameter, allows for a single fractional-order operator to interpolate between derivatives of all orders. This allows the user to directly control the parsimony, by fitting a specified number of fractional derivatives to data, without making assumptions on the orders of derivatives to be discovered. Therefore, building on a basic example of a fractional-order operator treated in \cite{RAISSI2017683}, we significantly extend the framework to treat other space-fractional differential operators and covariance kernels. The main problem that must be addressed is the efficient computation of the action of the unknown (fractional) linear operator $\mathcal{L}^{p_1,...,p_k}$ on covariance kernels. 

At the same time, fractional-order derivatives are far more than a tool to interpolate between integer-order and facilitate data-driven discovery of PDEs using continuous optimization techniques. The advances in the present article further improve the ability to discover fractional-order partial differential equations (FPDEs) from real-world data. It is now well understood that FPDEs have a profound connection with anomalous diffusions and systems driven by heavy-tailed processes (\cite{meerschaert_sikorskii}, \cite{metzler_klafter}). While such heavy-tailed data abounds in the fields of, e.g., hydrology, finance, and plasma physics, FPDEs are currently underutilized as tools to model macroscopic properties of such systems. This can be attributed to the analytic difficulties of deriving FPDE models; not only is there an additional parameter, but the involved formulas and nonlocal nature of fractional-order derivatives make them significantly less attractive for specialists in other fields to work with. 

Machine Learning is a natural tool for ameliorating this issue. As proof of this concept, we point to the work \cite{PANG2017694}, which employed a multi-fidelity Gaussian Process method to discover fractional-order of the fractional advection-dispersion equation governing underground tritium transport through a
heterogeneous medium at the Macrodispersion Experimental site at Columbus Air Force Base. This resulted in improved and highly efficient fitting of variable-order fractional equation to data. Along more theoretical lines, \cite{2017InvPr..33f5006G} explored the determination, using Gaussian processes, of the fractional-order of an elliptic problem involving a Spectral fractional Laplacian on a bounded domain with Neumann boundary condition. In addition, the authors also proved wellposedness of the inverse Bayesian formulation of this problem in the sense of \cite{stuart_2010}. Our work differs from \cite{PANG2017694} in that we do not repeatedly solve the forward problem, and from \cite{2017InvPr..33f5006G} in that we do not place a prior on the fractional order or on other parameters; rather, the parameters are inferred by placing a prior and training on the solution and right-hand side of the equation. This allows for more flexibility with regard to the form of the equation and the inclusion of additional parameters.

In the aforementioned Gaussian Process framework of \cite{RAISSI2017683}, it is possible to treat time-derivatives by considering the data and equation in space-time. Time may be treated as another dimension in the covariance kernel. {An alternative method is suggested by the work of
\cite{2018JCoPh.357..125R}, in which learning of evolution equations is based on the numerical differentiation of Gaussian processes. There, the data is given at different ``snap-shots'' in time which are used to discretize the time-derivative. In this way, even nonlinear time-dependent equations can be discovered using correspondence between nonlinear terms and specific linearizations of the discretized system.} The training is similar to that of \cite{RAISSI2017683}, and we extend this to fractional operators as well. 

This work is organized as follows. In section \ref{framework}, we review the Gaussian Process framework of \cite{RAISSI2017683} and \cite{2018JCoPh.357..125R}.
 In section \ref{kernels}, we review the Mat\'ern family of covariance kernels and space-fractional operators and present new formulas for space-fractional derivatives of such covariance kernels. Such covariance kernels can be more suited to rough data in certain applications. The inclusion of these new formulas, which can be efficiently treated using generalized Gauss-Laguerre quadrature, allows the discovery of various fractional equations in a unified way. In section \ref{examples} we present basic synthetic examples to illustrate the methodology, and in section 
\ref{interpolation} we apply the methodology to the discovery and interpolation of integer-order advection and diffusion using the fractional-order framework. This includes an interesting problem of interpolating two-term advection-diffusion using a single fractional-order term, and an exploration of user-selected parsimony. After reviewing the relation between $\alpha$-stable processes and fractional diffusion in section 
\ref{variable_order_matern} and studying a synthetic example, in section \ref{finance} we apply the methodology to the modeling of relative stock performance (Intel to S\&P 500) by $\alpha$-stable processes via an associated fractional-order diffusion equation. 

\section{The Gaussian Process Framework}\label{framework}
We review the framework developed by \cite{RAISSI2017683} for parametric learning of linear differential equations, of the form 
\begin{equation}
\mathcal{L}^{\bm{p}}_{\bm{x}} u = \mathcal{L}^{p_1,...,p_k}_{\bm{x}} u = f
\end{equation}
given data on $f$ and $u$.  Here, $\mathcal{L}^{p_1,...,p_k}_{\bm{x}}$ is a linear operator with unknown parameters $p_1,...p_k$. 
Here, and throughout the article, we use boldface characters (such as $\bm{x}$) to denote vector-valued variables, and capital boldface characters (such as $\bm{X}$) to denote data vectors. 
 
Assume $u(\bm{x})$ to be Gaussian process with mean $0$ and covariance function \\
$k_{uu}(\bm{x},\bm{x'};\bm{\theta})$ with hyperparameters $\bm{\theta}$:
\begin{equation}
\label{e:original_gp}
u(\bm{x}) \sim \mathcal{GP}(0, k_{uu}(\bm{x},\bm{x'};\bm{\theta})).
\end{equation}
We shall be vague about the form of the covariance kernel until Section \ref{kernels}; for now, it suffices to say that it describes how the correlation between the values of $u$ at two points $\bm{x}$ and $\bm{x'}$ falls off with $|\bm{x-x'}|$ or otherwise behaves with the two points, and that it must be a symmetric, positive semidefinite function \cite{rasmussen_williams}. 
Then, the linear transformation $f =\mathcal{L}^{\bm{p}}u$ of the Gaussian process $u$ implies a Gaussian Process for $f(\bm{x})$ {(see \cite{rasmussen_williams}, \S 9.4)},
\begin{equation}
f(\bm{x}) \sim \mathcal{GP}(0, k_{ff}(\bm{x},\bm{x'};\bm{\theta},\bm{p})),
\end{equation}
with covariance kernel
\begin{equation}\label{eq:kernelk}
k_{ff}(\bm{x},\bm{x'};\bm{\theta},\bm{p}) = \mathcal{L}_{\bm{x}}^{\bm{p}} \mathcal{L}^{\bm{p}}_{\bm{x'}} k_{uu}(\bm{x},\bm{x'};\bm{\theta}).
\end{equation}
Moreover, the covariance between $u(\bm{x})$ and $f(\bm{x}')$, and between $f(\bm{x})$ and $u(\bm{x}')$ is
\begin{align}\label{eq:kernelkuf}
\begin{split}
k_{uf}(\bm{x},\bm{x'};\bm{\theta},\bm{p}) &= \mathcal{L}^{\bm{p}}_{\bm{x'}} k_{uu}(\bm{x},\bm{x'};\theta) \\
k_{fu}(\bm{x},\bm{x'};\bm{\theta},\bm{p}) &= \mathcal{L}^{\bm{p}}_{\bm{x}} k_{uu}(\bm{x},\bm{x'};\theta),
\end{split}
\end{align}
respectively. By symmetry of $k_{uu}$, 
\begin{equation}
k_{fu}(\bm{x},\bm{x'};\bm{\theta},\bm{p}) = k_{uf}^T(\bm{x},\bm{x'};\bm{\theta},\bm{p}) \defeq k_{uf} (\bm{x'},\bm{x};\bm{\theta},\bm{p}).
\end{equation} 
The hyperparameters $(\bm{\theta},\bm{p})$ of the joint Gaussian Process
\begin{equation}\label{eq:joint_GP_uf}
\begin{bmatrix}
u(\bm{x}) \\ 
f(\bm{x})
\end{bmatrix}
\sim \mathcal{GP}\left(\bm{0}, \begin{bmatrix}
k_{uu}(\bm{x},\bm{x'};\bm{\theta}) & k_{uf}(\bm{x},\bm{x'};\bm{\theta},\bm{p})\\ 
k_{fu}(\bm{x},\bm{x'};\bm{\theta},\bm{p}) & k_{ff}(\bm{x},\bm{x'};\bm{\theta},\bm{p})
\end{bmatrix}
\right).
\end{equation}
are then learned by training on the data $\bm{Y}_u$ of $u$ given at points $\bm{X}_u$ and $\bm{Y}_f$ of $f$ given at points $\bm{X}_f$. This is done with a Quasi-Newton optimizer L-BFGS to minimize the negative log marginal likelihood (\cite{rasmussen_williams}):
\begin{align}
\label{NLML}
\begin{split}
\mathcal{NLML}(\bm{\theta},\bm{p},\sigma_{n_u}^2, \sigma_{n_f}^2) &= -\log p(\bm{Y}| \bm{\theta}, \bm{p}, \sigma_{n_u}^2, \sigma_{n_f}^2) \\
&= \frac{1}{2} \bm{Y}^{T}\bm{K}^{-1}\bm{Y} + \frac{1}{2}\log |\bm{K}| + \frac{N}{2} \log (2\pi),
\end{split}
\end{align}
where $\bm{Y} = \left[\begin{array}{c}
\bm{Y}_u \\ 
\bm{Y}_f
\end{array} \right]$, $p(\bm{Y} | \bm{\theta}, \bm{p},\sigma_{n_u}^2,\sigma_{n_f}^2) = \mathcal{N}\left(\bm{0}, \bm{K}\right)$, and $\bm{K}$ is given by
\begin{equation}
\label{covariance_matrix_1}
\bm{K} = \left[ \begin{array}{cc}
k_{uu}(\bm{X}_u,\bm{X}_u; \bm{\theta}) + \sigma_{n_u}^2 \bm{I}_{n_u} & k_{uf}(\bm{X}_u,\bm{X}_f;\bm{\theta},\bm{p})\\
k_{fu}(\bm{X}_f,\bm{X}_u;\bm{\theta},\bm{p}) & k_{ff}(\bm{X}_f,\bm{X}_f;\bm{\theta},\bm{p}) + \sigma_{n_f}^2 \bm{I}_{n_f}
\end{array}  \right].
\end{equation}
The additional noise parameters $\sigma_{n_u}^2$ and $\sigma_{n_f}^2$ are included to learn {uncorrelated} noise in the data; their inclusion above corresponds to the assumption that 
\begin{align}
\begin{split}
\bm{Y}_u &= u(\bm{X}_u) + \bm{\epsilon}_u \\
\bm{Y}_f &= f(\bm{X}_f) + \bm{\epsilon}_f \\
\end{split}
\end{align}
with $\bm{\epsilon}_u \sim \mathcal{N}(\bm{0},\sigma_{n_u}^2 \bm{I}_{n_u})$ and independently $\bm{\epsilon}_f \sim \mathcal{N}(\bm{0},\sigma_{n_f}^2 \bm{I}_{n_f})$. 

Next we review the time-stepping Gaussian Process method of \cite{2018JCoPh.357..125R} for learning linear (in our case) equations of the form
\begin{eqnarray}\label{eq:PDE}
u_t + \mathcal{L}^{p_1,...,p_k} u = 0,\ x \in \mathbb{R}^d, \ t\in[0,T].
\end{eqnarray}
For our purposes, we consider two ``snapshots'' $\{\bm{x}^{n-1}, \bm{u}^{n-1}\}$ and $\{\bm{x}^{n}, \bm{u}^{n}\}$ of the system at two times $t^{n-1}$ and $t^n$, respectively, such that 
\begin{equation}
t^n - t^{n-1} = \Delta t \ll 1.  
\end{equation}
We perform, in the case of two snapshots, a backward Euler discretization
\begin{equation}\label{eq:BackwardEuler}
u^{n} + \Delta t \mathcal{L}^{p_1,...,p_k} u^n = u^{n-1}.
\end{equation}
Then we assume a Gaussian Process, but for $u^n$:
\begin{equation}\label{eq:Prior}
u^n(\bm{x}) \sim \mathcal{GP}(0, k(\bm{x},\bm{x'},\bm{\theta})).
\end{equation}
As before, the linearity of $\mathcal{L}^{p_1,...,p_k}$ leads to a Gaussian Process for 
$u^{n-1}$. We obtain the joint Gaussian process
\begin{equation}\label{eq:HPM}
\begin{bmatrix}
u^{n} (\bm{x}) \\ 
u^{n-1} (\bm{x})
\end{bmatrix}
\sim \mathcal{GP}\left(\bm{0}, \begin{bmatrix}
k^{n,n}  (\bm{x},\bm{x'};\bm{\theta})  & k^{n,n-1}  (\bm{x},\bm{x'};\bm{\theta},\bm{p}) \\ 
k^{n-1,n}  (\bm{x},\bm{x'};\bm{\theta},\bm{p})  & k^{n-1,n-1}  (\bm{x},\bm{x'};\bm{\theta},\bm{p})
\end{bmatrix}
\right).
\end{equation}
where, denoting the identity operator by $\text{Id}$,
\begin{align}
\label{time_dependent_kernels}
&k^{n,n} = k, &&k^{n,n-1} = (\text{Id}+\Delta t\mathcal{L}_{\bm{x'}}^{\bm{p}}) k,\\ \nonumber
&k^{n-1,n} = (\text{Id} +\Delta t\mathcal{L}_{\bm{x}}^{\bm{p}}) k, 
&&k^{n-1,n-1} = (\text{Id} +\Delta t\mathcal{L}_{\bm{x}}^{\bm{p}}) (\text{Id}+\Delta t\mathcal{L}_{\bm{x'}}^{\bm{p}}) k.
\end{align}
Equation \eqref{eq:HPM} can be compared to equation \eqref{eq:joint_GP_uf}, and equation \eqref{time_dependent_kernels} to \eqref{eq:kernelk} and \eqref{eq:kernelkuf}. The set-ups are very similar, and again, 
$\bm{p}$ has been merged into the hypermarameters of this joint Gaussian Process. 
Given data at spacial points  $\bm{X}^n$ and $\bm{X}^{n-1}$ for the functions $u^n$ and $u^{n-1}$, represented by the vectors $\bm{U}^n$ and $\bm{U}^{n-1}$, respectively, the new hyperparameters $(\bm{\theta}, \bm{p})$ are trained by employing the same Quasi-Newton optimizer L-BFGS as before to minimize the $\mathcal{NLML}$ given by equation \eqref{NLML}. In this case,
 $\bm{Y} = \begin{bmatrix}
\bm{U}^n \\ 
\bm{U}^{n-1}
\end{bmatrix}
$, $p(\bm{y} | \bm{\theta}, {\bm{p}}, \sigma^2) = \mathcal{N}\left(\bm{0}, \bm{K}\right)$, and $\bm{K}$ is given by
\begin{equation}
\label{time_dependent_kernels_data}
\bm{K} = \begin{bmatrix}
k^{n,n}(\bm{X}^{n},\bm{X}^{n}) & k^{n,n-1}(\bm{X}^{n},\bm{X}^{n-1})\\ 
k^{n-1,n}(\bm{X}^{n-1},\bm{X}^{n}) & k^{n-1,n-1}(\bm{X}^{n-1},\bm{X}^{n-1})
\end{bmatrix} + \sigma_n^2 \bm{I}.
\end{equation}
Here, $\sigma_n^2$ is an additional parameter to learn noise in the data, under the assumption
\begin{align}
\begin{split}
\bm{u}^n &= u^n(\bm{X}^n) + \bm{\epsilon}^n \\
\bm{u}^{n-1} &= u^{n-1}(\bm{X}^{n-1}) + \bm{\epsilon}^{n-1}
\end{split}
\end{align}
with $\bm{\epsilon}^n \sim \mathcal{N}(0, \sigma_n^2 \bm{I})$ and $\bm{\epsilon}^{n-1} \sim \mathcal{N}(0, \sigma_n^2 \bm{I})$ being independent.

\section{Fractional Derivatives of Covariance Kernels}\label{kernels}
Many properties of a Gaussian Process are determined by the choice of covariance kernel $k(\bm{x},\bm{x'})$. In particular, the covariance kernel encodes an assumption about the smoothness of the field that being interpolated. Stein  \cite{stein} writes ``...properties of spatial interpolants depend
strongly on the local behavior of the random field. In practice, this local
behavior is not known and must be estimated from the same data that
will be used to do the interpolation. This state of affairs strongly suggests
that it is critical to select models for the covariance structures that include
at least one member whose local behavior accurately reflects the actual
local behavior of the spatially varying quantity under study''. Mat\'ern kernels $M_\nu$ (defined below), a family of stationary kernels which includes the exponential kernel for $\nu = 1/2$, and the squared-exponential kernel in the limit 
$\nu \rightarrow \infty$, have been widely used for this reason.  
A Gaussian Process with Mat\'ern covariance kernel $M_\nu$ is $n$-times mean square differentiable for $n > \nu$. 
We have developed a computational methodology that allows for Mat\'ern kernels of arbitrary real order $\nu > 0$ to be used in our Gaussian Process, namely in \eqref{e:original_gp} and \eqref{eq:Prior}. In fact, the parameter $\nu$ itself can be treated and optimized as a hyperparameter of the Gaussian Process, just as the equation parameters were in Section \ref{framework}. We employ such an algorithm to explore the effect of the parameter $\nu$ when working with rough time series histogram data in section \ref{variable_order_matern}. 

{The main problem that arises when using fractional operators is the computation of their action on kernels such as the Mat\'ern class, as required by equations \eqref{eq:kernelk} and \eqref{eq:kernelkuf} for the time-independent case and \eqref{time_dependent_kernels} for the time-dependent case. This cannot be done analytically, and requires a numerical approach, in contrast to the works \cite{RAISSI2017683} and \cite{2018JCoPh.357..125R} where (standard) differential operators applied to kernels were obtained symbolically in closed form using Mathematica.  Moreover, unlike standard derivatives, fractional derivatives, whether in $\mathbb{R}^d$, $\mathbb{R}^+$, or on bounded subsets, are nonlocal operators typically defined by singular integrals or eigenfunction expansions that are difficult and expensive to discretize \cite{2018arXiv180109767L},  \cite{meerschaert_sikorskii}.} However, space-fractional derivatives in $\mathbb{R}^d$ enjoy representations as Fourier multiplier operators. In other words, they are equivalent to multiplication by a function $m(\bm{\xi})$ in frequency space.
This representation suggests a computational method that avoids any singular integral
operators or the solution of extension problems in
$\mathbb{R}^{d+1}$.
The downside to Fourier methods is that, if used to compute the
fractional derivative of a function $u$ on  $\mathbb{R}^{d}$, they may
require quadrature of a $2d$ (forward
and inverse) Fourier integral. Thus, if one wishes to compute the
fractional derivative $\mathcal{L}$ of a covariance kernel
$k(\vec{x},\vec{y})$, as in \eqref{eq:kernelk}, \eqref{eq:kernelkuf}, or \eqref{time_dependent_kernels}, this may entail $2d$ quadrature for $
\mathcal{L}_{\bm{x}} k$
and $\mathcal{L}_{\bm{y}} k$, and $4d$ quadrature for $
\mathcal{L}_{\bm{x}}\mathcal{L}_{\bm{y}} k$.
These dimensions for quadrature can be cut in half provided the
(forward) Fourier transforms of these kernels were known analytically.
Moreover, if the covariance kernel $k_{uu}$ is stationary,  we see in 
Theorem \ref{kernel_theorem} below that $\mathcal{L}_{\bm{x}}\mathcal{L}_{\bm{y}} k$ can further 
be reduced from a $2d$-dimensional integral to $d$-dimensional one.

Thus, the entire problem of kernel computation is reduced to
$d$-dimensional quadrature if the following three conditions are
satisfied:

\begin{enumerate}
\item
\emph{The spacial differential operator $\mathcal{L}$ is a Fourier
multiplier operator}:
\begin{equation}
\mathcal{F} \{ \mathcal{L}f \}(\vec{\xi}) = m(\vec{\xi}) \cdot
\mathcal{F} \{ f \}(\vec{\xi}).
\end{equation}
This is true for a variety of fractional space derivatives:
\begin{align}
\begin{split}
\label{derivatives_defs}
\text{Fractional Laplacian
: }&\mathcal{F} \{ (-\Delta)^{\alpha/2} f
\}(\vec{\xi}) = |\vec{\xi}|^\alpha \mathcal{F} \{ f \}(\vec{\xi}) \\
\text{Left-sided Riemann-Louiville: }&\mathcal{F} \{_{-\infty}
^{RL} D^\alpha_{\bm{x}} f \}(\vec{\xi}) = (-i\vec{\xi})^\alpha \mathcal{F} \{ f
\}(\vec{\xi}) \\
\text{Right-sided Riemann-Louiville: }&\mathcal{F} \{_{\bm{x}}
^{RL} D^\alpha_{\infty} f \}(\vec{\xi}) = (i\vec{\xi})^\alpha \mathcal{F} \{
f \}(\vec{\xi}).
\end{split}
\end{align}
Here, and throughout this article, we use the Fourier transform convention
\begin{equation}
\mathcal{F} f = \frac{1}{(2\pi)^{d/2}} \int_{\mathbb{R}^d} e^{-i \bm{\xi} \cdot \bm{x} } f(\bm{x}) d\bm{x},
\quad
\mathcal{F}^{-1} \hat{f} = \frac{1}{(2\pi)^{d/2}} \int_{\mathbb{R}^d} e^{i \bm{\xi} \cdot \bm{x} } \hat{f}(\bm{\xi}) d\bm{\xi}
\end{equation}
\item
\emph{The covariance kernel $k$ is stationary}:
\begin{equation}
k(\vec{x},\vec{y}) = K(\vec{x}-\vec{y}).
\end{equation}
This is true of the squared-exponential kernel in one-dimension
\begin{equation}
G(\sigma, \theta; x,y) = \sigma^2
\exp \left(-\frac{1}{2}
\frac{({x}-{y})^2}{\theta^2} \right)
\end{equation}
as well as frequently used multivariate squared-exponential kernels,
formed by multiplication
\begin{equation}
G^\times(\sigma, \theta_1, \theta_2, ..., \theta_d; \bm{x}, \bm{y}) =
\sigma^2 \prod_{i=1}^d
\exp \left(-\frac{1}{2}
\frac{({x}_i-{y}_i)^2}{\theta_i^2} \right)
\end{equation}
or addition
\begin{equation}
G^+ (\sigma, \theta_1, \theta_2, ..., \theta_d; \bm{x}, \bm{y}) =
\sigma^2 \sum_{i=1}^d
\exp \left(-\frac{1}{2}
\frac{({x}_i-y_i)^2}{\theta_i^2} \right)
\end{equation}
of the one-dimensional kernel.
The same is true for the Mat\'ern kernels $M_\nu$, which have one-dimensional form
\begin{equation}
\label{matern_1d}
M_\nu(\sigma, \theta; x,y) = \frac{\sigma^2 2^{1-\nu}}{\Gamma(\nu)}
\left( \frac{\sqrt{2\nu}}{\theta}({x}-y)\right)^\nu
K_{\nu}\left(\frac{\sqrt{2\nu}}{\theta}({x}-y)\right)
\end{equation}
and the corresponding multidimensional kernels
\begin{align}
\label{multi_d_matern_kernels}
\begin{split}
M_{\nu_1, ..., \nu_d}^\times(\bm{x,y}) &= \prod_{i=1}^d M_{\nu_i}({x}_i-y_i), \\
M_{\nu_1, ..., \nu_d}^+(\bm{x,y}) &= \sum_{i=1}^d M_{\nu_i}({x}_i-y_i).
\end{split}
\end{align}
The notation ${K_{\nu}}$ refers to the modified Bessel
function, which is potentially confusing, but will not be an issue as we 
will focus on Fourier representation of
${M_\nu}$ in what follows.

\item
\emph{The (forward) Fourier transform $\hat{K}(\vec{\xi}) =
\mathcal{F}\{K\}(\vec{\xi})$ of the stationary covariance kernel $K$
is known analytically.} This is satisfied by the squared exponential kernel
$G(\sigma, \theta; x)$ and the Mat\'ern
kernel\footnote{In the machine learning literature, authors such as \cite{rasmussen_williams}
describe this Fourier transform as the \emph{spectral density} in the context of 
Bochner's theorem on stationary kernels, and write it in the equivalent form, up to Fourier transform convention:
$
 \hat{M}_\nu({\xi}) = \sigma^2 \frac{\sqrt{2}\Gamma(\nu+1/2)(2\nu)^\nu}
{\Gamma(\nu)\theta^{2\nu}}
\left(
\frac{2\nu}{\theta^2} + \xi^2
\right)^{-(\nu+1/2)}
$.}
$M_\nu(\sigma, \theta;x)$:
\begin{align}
\begin{split}
\label{kernel_fourier_transforms}
\text{Squared-exponential kernel: }&\mathcal{F}\{ G \}(\xi) =
 \theta
\sigma^2 e^{-\frac{\theta^2}{2} \xi^2}
\\
\text{Mat\'ern kernel: }&\mathcal{F}\{ M_\nu \}(\xi) =
\frac{\theta \sigma^2 \Gamma(\nu+1/2) }
{\sqrt{\nu} \Gamma(\nu) }
\left(
1+ \frac{\theta^2 \xi^2}{2\nu}
\right)^{-(\nu+1/2)}
\end{split}
\end{align}
The same is true for the
multidimensional kernels
$G^+$ and $M_\nu^+$ by linearity of the Fourier transform,
and for 
$G^\times$ and
$M_\nu^\times$
by Fubini's theorem. 
\end{enumerate}

\begin{theorem}
\label{kernel_theorem}
Suppose conditions (1)-(3) on the covariance kernel $k$ and the operator $\mathcal{L}$ and are satisfied. 
Then the fractional derivatives of the covariance kernel 
$\mathcal{L}_{\bm{x}} k $, $\mathcal{L}_{\bm{y}} k = \left[\mathcal{L}_{\bm{x}} k\right]^T$, and 
$\mathcal{L}_{\bm{y}} \mathcal{L}_{\bm{x}} k$ can be computed by $d$-dimensional 
integrals
\begin{equation}
\label{kernel_theorem_equations}
\begin{cases}
\begin{aligned}
\mathcal{L}_{\bm{x}} k &=
\frac{1}{\sqrt{2\pi}}
\int_{\mathbb{R}^d}
e^{i\langle \bm{x-y}, \bm{\xi} \rangle}
m(\bm{\xi})
\hat{K}(\bm{\xi})
d\bm{\xi} \\
\mathcal{L}_{\bm{y}} \mathcal{L}_{\bm{x}} k
&=
\frac{1}{\sqrt{2\pi}}
\int_{\mathbb{R}^d}
e^{i\langle \bm{x-y} , \bm{\xi} \rangle}
m(\bm{\xi})m(-\bm{\xi})
\hat{K}(\bm{\xi})
d\bm{\xi}.
\end{aligned}
\end{cases}
\end{equation}
\end{theorem}
For a proof of this theorem, see Appendix \ref{appendix}.

{Provided the integrals \eqref{kernel_theorem_equations} can be computed numerically at the locations of the data, they can be used to build the kernel matrix $\bm{K}$ and evaluate the objective function $\mathcal{NLML}$ given by \eqref{NLML}. In the Quasi-Newton L-BFGS method (discussed in Section \ref{framework}) that is used to train the extended hyperparameters $\bm{\theta} = [\theta_i]$ implicit in $\bm{K}$, we supply the gradient of the $\mathcal{NLML}$, the components of which are given by 
\begin{equation}
\frac{\partial \mathcal{NLML}}{\partial {\theta}_i}
=
\frac{1}{2}\text{Tr}\left(\bm{K}^{-1} \frac{\partial \bm{K}}{\partial
\theta_i}\right)
-
\frac{1}{2}
\bm{Y}^T \bm{K}^{-1} \frac{\partial \bm{K}}{\partial \theta_i} \bm{K}^{-1} \bm{Y}.
\end{equation}
See \cite{rasmussen_williams}, \S 5.4. To obtain ${\partial \bm{K}}/{\partial\theta_i}$, we note in \eqref{kernel_theorem_equations} that the derivative ${\partial}/{\partial\theta_i}$ may be passed into the integrand and through the complex exponential factor. The resulting derivative of the product of the multiplier(s) $m$, which contains the equation parameters, and $\hat{K}$, which contains the original kernel parameters, can be obtained symbolically as a closed-form expression.  The same numerical procedure is used to evaluate the resulting integrals as for \eqref{kernel_theorem_equations}. The Hessian is not supplied in closed form and is approximated from evaluations of the gradient. 
}

When training a Gaussian process, it is advantageous to \emph{standardize} the data \cite{data_mining} so that     
 $\bm{x}, \bm{y}, \theta \sim \mathcal{O}(1)$ when possible. In the framework discussed here, this reduces the difficulty of computing the kernel functions in \eqref{kernel_theorem_equations} by restricting the frequency and support of the integrands. When necessary, standardization for the applications considered here can be performed by rescaling the values and positions of the data point  by appropriate constants. Once the differential equation is learned for the scaled solution $u_{\text{scaled}} = 
Au(Bx)$, the true coefficients can be obtained via inverse scaling. This is discussed in detail for an example in Section \ref{finance}. {During training, we expect convergence to a local minimum of the $\mathcal{NLML}$, and we have not found the optimal $\bm{\theta}$ to depend significantly on the initial guess in our examples, but there is no guarantee of this. Uniqueness, stability, and convergence remain important open questions.}

Although numerical calculation of the above integrals can be performed using
Gauss-Hermite quadrature, this is not optimal as the fractional-order monomial
$m(\bm{\xi})$ is not smooth at the origin. To obtain faster convergence with the number
of quadrature points, a superior choice is generalized Gauss-Laguerre quadrature, involving
a weight function of the form $x^{\alpha_{\text{gGL}}} e^{-x}$ for $\alpha_{\text{gGL}} > -1$:
\begin{equation}
\int_0^\infty f(x) dx = \int_0^\infty 
x^{\alpha_{\text{gGL}}} e^{-x} \left[ e^{x} x^{-\alpha_{\text{gGL}}}  f(x) \right] dx \
{  \approx} 
 \sum_i^n  w_i e^{x_i} x_i^{-\alpha_{\text{gGL}}} f(x_i). 
\end{equation}
{Here, $w_i$ are the Gauss-Laguerre  weights, and $x_i$ the nodes.}  
In practice, it is essential for $\alpha_{\text{gGL}}$ to match the fractional part of the power of 
the monomial in the integrand $f$, as the remainder yields a smooth function.
We use the  Golub-Welsch algorithm to find
the nodes, but compute the weights by evaluating the generalized Gauss-Laguerre polynomial
at these nodes for higher relative accuracy.

We describe two examples and discuss the convergence of the numerical Gauss-Laguerre quadrature in each one. 
First, consider the left-sided Riemann-Louiville derivative $\mathcal{L}_x = _{-\infty}^{RL} D^\alpha_{{x}}$
in one dimension, which involves $m(\xi) = (-i\xi)^\alpha$ in the above formulas. 
Owing to the symmetry of $\hat{K}(\xi)$, one can write
\begin{align}
_{-\infty} ^{RL} D^\alpha_{{x}} k &=
\frac{1}{\sqrt{2\pi}}
\int_{-\infty}^\infty
e^{i ({x-y}, \xi)}
(-i\xi)^\alpha
\hat{K}({\xi})
d{\xi} \\
&=
\frac{1}{\sqrt{2\pi}}
\left[ \int_0^\infty
e^{i ({x-y}, \xi)}
(-i\xi)^\alpha
\hat{K}({\xi})
d{\xi}
+
 \int_{-\infty}^0
e^{i ({x-y}, \xi)}
(-i\xi)^\alpha
\hat{K}({\xi})
d{\xi} 
\right] \\
&=
\frac{1}{\sqrt{2\pi}}
\left[ \int_0^\infty
e^{i ({x-y}, \xi)}
(-i\xi)^\alpha
\hat{K}({\xi})
d{\xi}
+
 \int_0^\infty
\overline{
e^{i ({x-y}, \xi)}
(-i\xi)^\alpha
}
\hat{K}({\xi})
d{\xi} 
\right] \\
&=
\frac{2}{\sqrt{2\pi}}
\int_0^\infty
\text{Re}\left[
e^{i ({x-y}, \xi)}
(-i\xi)^\alpha
\hat{K}({\xi})
\right]
d{\xi}
\end{align}
Similarly,
\begin{equation}
_{-\infty} ^{RL} D^\alpha_{{y}} \left[ _{-\infty} ^{RL} D^\alpha_{{x}} k \right]
=
\frac{2}{\sqrt{2\pi}}
\int_0^\infty
\text{Re}\left[
e^{i ({x-y}, \xi)}
(-i\xi)^\alpha(i\xi)^\alpha
\hat{K}({\xi})
\right]
d{\xi}.
\end{equation}
These integrals call for generalized Gauss-Laguerre quadrature to be performed with $\alpha_{\text{gGL}} = \alpha$ for 
$_{-\infty}^{RL} D^{\alpha}_{x} k$ and $\alpha_{\text{gGL}} = 2\alpha$ for 
${_{-\infty}^{RL} D^{\alpha}_{y} } \left[ _{-\infty}^{RL} D^{\alpha}_{x} k \right]$. Using the Mat\'ern kernel with $\nu = 5/2$ as an example, 
the convergence of the error with the number of quadrature points is shown in Table \ref{1d-error}. The MATLAB \texttt{integral} function is used for reference when computing the error. 
The setup for working with the right-sided Riemann-Louiville derivative, or for the one-dimensional fractional Laplacian,
is entirely similar.

Next we consider the fractional Laplacian $\mathcal{L} = (-\Delta)^{\alpha/2}$ in two dimensions. 
This involves the multiplier $m(\bm{\xi}) = |\bm{\xi}|^{\alpha} = |\xi_1^2 + \xi_2^2|^{\alpha/2}$. 
We transform the integrals into polar coordinates: 
\begin{align}
(-\Delta_x)^{\alpha/2} k &=
\frac{1}{\sqrt{2\pi}}
\int_0^{2\pi}
\int_0^\infty
e^{i\langle \bm{x} - \bm{y}, (r\cos\theta, r\sin\theta) \rangle}
%e^{i[(y_1-x_1)r\cos\theta + (y_2 - x_2)r\sin\theta]}
r^\alpha
\hat{K}(r\cos\theta, r\sin\theta)
r dr d\theta. \\
(-\Delta_y)^{\alpha/2} (-\Delta_x)^{\alpha/2} k
&=
\frac{1}{\sqrt{2\pi}}
\int_0^{2\pi}
\int_0^\infty
e^{i\langle \bm{x} - \bm{y}, (r\cos\theta, r\sin\theta) \rangle}
%e^{i[(y_1-x_1)r\cos\theta + (y_2 - x_2)r\sin\theta]}
r^{2\alpha}
\hat{K}(r\cos\theta, r\sin\theta)
r dr d\theta.
\end{align}
The quadrature of these integrals is performed using the trapezoid
rule in $\theta$ and generalized Gauss-Laguerre quadrature in $r$, using
$\alpha_{\text{gGL}} = \alpha+1$ for $(-\Delta_x)^{\alpha/2} k$ and
$\alpha_{\text{gGL}} = 2\alpha+1$ for $(-\Delta_y)^{\alpha/2} (-\Delta_x)^{\alpha/2} k$. 
Using the product multivariate Mat\'ern kernel $M_{\frac{5}{2}} M_{\frac{7}{2}}$ as an example,
convergence with respect to the number of quadrature points in $r$
 is show in in Table \ref{2d-error}, where 64 quadrature points in $\theta$ and 8, 16, 32, 64
quadrature points in $r$ are used. 
Reference answers were generated by nesting the MATLAB \texttt{integral} function.

The examples in the following sections are built by defining the Matern kernels 
in Mathematica and performing derivatives with respect to the hyperparameters symbolically. 
The expressions are ported into MATLAB code where Gauss-Laguerre quadrature with appropriate
$\alpha_{\text{gGL}}$ is used to compute the $\mathcal{NLML}$ \eqref{NLML} as well as the derivatives of the 
$\mathcal{NLML}$ with respect to the parameters. For our purposes, 64 quadrature points in one dimension and 64x64 quadrature points (as described above) in two dimensions is sufficient. 

\setlength{\tabcolsep}{0.15em} % for the horizontal padding
{\renewcommand{\arraystretch}{1.10}% for the vertical padding
\newcolumntype{x}[1]{>{\centering\arraybackslash\hspace{0pt}}p{#1}}

% \usepackage{multirow}
\begin{table}[]
\small
\caption{\small$L^{\infty}$ error in computing the Mat\'ern kernel $M_{\frac{5}{2}}(x-y)$ and the kernel blocks  
$\mathcal{L}_x M_{\frac{5}{2}}(x-y)$ and $\mathcal{L}_x \mathcal{L}_y M_{\frac{5}{2}}(x-y)$, with
$\mathcal{L}_x = ^{RL}_x D^{1/2}_\infty $,
on $[-1,1]$ using generalized Gauss-Laguerre quadrature.}
\label{1d-error}
\begin{tabular}{|x{1.05in}|x{0.25in}|x{0.75in}|x{0.75in}|x{0.75in}|x{0.75in}|}
\hline
\multirow{2}{*}{\begin{tabular}[c]{@{}c@{}}Function\\ $\left( \mathcal{L}_x = ^{RL}_x D^{1/2}_\infty\right)$\end{tabular}} & \multirow{2}{*}{$\theta^2$} & \multicolumn{4}{c|}{\begin{tabular}[c]{@{}c@{}}$L^\infty[-1,1]$-norm of error as function of $(x-y)$\\ @ number of quadrature points\end{tabular}} \\ \cline{2-6} 
                                                                                                                           &                             &  8                                               & 16                                              & 32                     & 64                       \\ \hline

\multirow{4}{*}{$M_{\frac{5}{2}}(x-y)$}                                                                                    & $\frac{1}{10}$              & 1.190e-02                                       & 3.468e-04                                       & 2.917e-06              &2.876e-07               \\ \cline{2-6} 
                                                                                                                           & 1                           & 5.126e-03                                       & 8.029e-06                                       & 3.741e-07              &3.204e-07               \\ \cline{2-6} 
                                                                                                                           & 10                          & 8.163e-02                                       & 1.446e-02                                       & 2.378e-04              &4.188e-06                 \\ \hline
\multirow{4}{*}{$\mathcal{L}_x M_{\frac{5}{2}}(x-y)$}                                                                      & $\frac{1}{10}$              & 4.752e-02                                       & 1.991e-03                                       & 2.822e-05              &3.315e-06                \\ \cline{2-6} 
                                                                                                                           & 1                           & 6.179e-03                                       & 9.528e-05                                       & 2.524e-07              &1.634e-07                \\ \cline{2-6} 
                                                                                                                           & 10                          & 6.937e-02                                       & 9.915e-03                                       & 3.456e-04              &2.828e-06                \\ \hline
\multirow{4}{*}{$\mathcal{L}_x \mathcal{L}_y M_{\frac{5}{2}}(x-y)$}                                                        & $\frac{1}{10}$              & 1.624e-01                                       & 8.980e-03                                        & 2.277e-04              &3.449e-06              \\ \cline{2-6} 
                                                                                                                           & 1                           & 1.006e-03                                        & 1.515e-04                                       & 9.092e-07              &9.642e-07               \\ \cline{2-6} 
                                                                                                                           & 10                          & 6.571e-02                                       & 4.774e-03                                       & 3.370e-04              &1.217e-06                \\ \hline
\end{tabular}
\end{table}

% \usepackage{multirow}
\begin{table}[]
\small
\caption{\small$L^{\infty}$ error in computing the Mat\'ern kernel $M_{\frac{5}{2}}M_{\frac{7}{2}}(\bm{x-y})$ and the kernel blocks  
$\mathcal{L}_{\bm{x}} M_{\frac{5}{2}}M_{\frac{7}{2}}(\bm{x-y})$ and $\mathcal{L}_{\bm{x}} \mathcal{L}_{\bm{y}} 
M_{\frac{5}{2}}M_{\frac{7}{2}}(\bm{x-y})$ on the square $[-1,1]^2$. The same correlation parameter $\theta$ is used for 
both kernels. Quadrature is performed generalized Gauss-Laguerre quadrature in the polar variable $r$ with 8, 16, 32, and 64 quadrature points, with a fixed number of 64 trapezoid rule quadrature points in $\theta$.}
\label{2d-error}
\begin{tabular}{|x{1.05in}|x{0.25in}|x{0.75in}|x{0.75in}|x{0.75in}|x{0.75in}|}
\hline
\multirow{2}{*}{\begin{tabular}[c]{@{}c@{}}Function\\ $\left(\mathcal{L}_x = (-\Delta_x)^{\alpha/2}\right)$\end{tabular}} & \multirow{2}{*}{$\theta^2$} & \multicolumn{4}{c|}{\begin{tabular}[c]{@{}c@{}}$L^\infty[-1,1]^2$-norm of error as function of $(\bm{x-y})$\\ @ number of quadrature points\end{tabular}} \\ \cline{3-6} 
                                                                                                              &                             & 8                                                 & 16                                                & 32                 &64                     \\ \hline
\multirow{4}{*}{$M_{\frac{5}{2}}M_{\frac{7}{2}}$}                                                             & $\frac{1}{10}$              & 5.406e-02                                         & 6.201e-04                                         & 8.248e-06                     &9.939e-05               \\ \cline{2-6} 
                                                                                                              & 1                           & 1.368e-02                                         & 3.611e-04                                         & 2.659e-06                     &2.955e-06              \\ \cline{2-6} 
                                                                                                              & 10                          & 7.955e-01                                         & 8.385e-02                                         & 3.873e-03                     &2.006e-05             \\ \hline
\multirow{4}{*}{$\mathcal{L}_x \left[ M_{\frac{5}{2}} M_{\frac{7}{2}} \right] $}                              & $\frac{1}{10}$              & 2.025e-01                                         & 3.922e-03                                         & 6.705e-05                     &3.842e-05            \\ \cline{2-6} 
                                                                                                              & 1                           & 2.769e-02                                         & 9.611e-04                                         & 8.921e-06                     &8.977e-06             \\ \cline{2-6} 
                                                                                                              & 10                          & 3.717e-01                                         & 2.013e-02                                         & 3.340e-03                     &1.515e-05             \\ \hline
\multirow{4}{*}{$\mathcal{L}_x \mathcal{L}_y \left[M_{\frac{5}{2}}M_{\frac{7}{2}}\right]$}                    & $\frac{1}{10}$              & 4.759e-01                                         & 2.644e-02                                         & 5.154e-04                     &7.347e-05            \\ \cline{2-6} 
                                                                                                              & 1                           & 8.117e-02                                         & 1.449e-03                                         & 8.740e-06                    &9.848e-06              \\ \cline{2-6} 
                                                                                                              & 10                          & 8.713e-02                                         & 4.553e-02                                         & 1.535e-03                  &8.209e-06                 \\ \hline
\end{tabular}
\end{table}

\FloatBarrier

\section{A Basic Example}\label{examples}

In this section, we will illustrate the methodology to discover the parameters $C$ and $\alpha$ in the fractional elliptic equation
\begin{equation}
C(-\Delta)^{\alpha/2} u = f
\end{equation}
in one and two space dimensions from data on $u$ and $f$. 

Following the time-independent framework, this means that we must optimize the negative log marginal likelihood \eqref{NLML}, where in the covariance matrix \eqref{covariance_matrix_1}, the kernels are given by the integral formulas \eqref{kernel_theorem_equations}. In the latter formulas, the multiplier $m$ and the Fourier transform $\hat{K}$ of the stationary prior kernel must be specified; the multiplier $m$ corresponding to the operator $C(-\Delta)^{\alpha/2}$ is $m(\xi) = C|\xi|^\alpha$ in one dimension, and $
m(\xi_1, \xi_2) = C|\xi_1^2 + \xi_2^2|^{\alpha/2}  
$
in two dimensions.
We choose to use the Mat\'ern kernel $k_{uu} = K = M_{\nu}$, given by \eqref{matern_1d} in one dimension, with a tensor product \eqref{multi_d_matern_kernels} of Mat\'ern kernels $M_{\nu_1, \nu_2}^\times$ in two dimensions. Thus, by equations \eqref{kernel_fourier_transforms},
$
\hat{K} = \hat{M}_\nu(\xi)
\text{ in one dimension, and }
\hat{K} = 
\hat{M}_{\nu_1}(\xi_1)
\hat{M}_{\nu_1}(\xi_2)
\text{ in two dimensions}. 
$
This completes the description of the covariance kernel.

In the one-dimensional case the solution/RHS pair
\begin{equation}
u = e^{-x^2}, \quad f = \frac{C}{\sqrt{2\pi}} \int_{-\infty}^{\infty}
e^{ix\xi} |\xi|^{\alpha}\frac{e^{-\xi^2/4}}{\sqrt{2}} d\xi
\end{equation}
is used.  Two data sets are generated for two experiments. In both experiments, we use the above solution/RHS pair to 
generate data with $\text{ exact } C = 1.25, \quad \alpha = \sqrt{2} \approx 1.4142$. The Mat\'ern kernel with fixed $\nu = 11/2$ is used to discover the parameters, 
and the initial parameters for the optimization are taken to be $\theta = 1, \sigma = 1, \alpha = 1.0 , C = 0.5$. 
In the first experiment, 7 data points at $\bm{X}_u$ for $u$ and  11 data points $\bm{X}_f$ for $f$ are generated via latin hypercube sampling. No noise is added.
In the second experiment, 20 data points for each of $u$ and $f$ are generated in the same way, but normal random noise 
of standard deviation 0.1 for $u$ and 0.2 for $f$ is added to the data.

The GP regression for the first experiment is shown in Figure \ref{ex1a}. The equation parameters recovered are 
$\alpha = 1.40158$ and $C = 1.25840$, which are within 1\% of the true values. The GP regression for the second
experiment is shown in Figure \ref{ex1b}. There, we have also plotted the twice the standard deviation of the Gaussian process 
plus twice the standard deviation of the learned noise, $2\sigma_{n_u}$ and $2\sigma_{n_f}$ (in the previous experiment, these 
learned parameters were miniscule). The parameters learned in the second
experiment are $\alpha = 1.51151$ and $C = 1.18099$, which are within 7\% and 6\% of the true values, respectively. 

In the two dimensional example, the exact solution pair used to generate data is now
\begin{equation}
u = e^{-x_1^2-x_2^2}, \quad f = \frac{C}{{2\pi}} \int_{-\infty}^{\infty} \int_{-\infty}^{\infty}
e^{ix_1\xi_1} e^{ix_2\xi_2} |\xi_1^2 + \xi_2^2|^{\alpha/2 }\frac{e^{-\xi_1^2/4-\xi_2^2/4}}{2} d\xi_1 d\xi_2
\end{equation}
with 
$
\text{ exact } C = 1, \quad \alpha = \sqrt{3} \approx 1.7321.
$
Again, the Mat\'ern parameter $\nu = \nu_1 = \nu_2 = 11/2$ is used. The initial parameters for the optimization are 
$\sigma=1, \theta_1 = 1, \theta_2 = 1, \alpha = 2 , C = 1.5$. We take 40 data points for each of $u$ and $f$, generated using latin hypercube sampling on $[-2,2]\times[-2,2]$. The parameters $\alpha = \sqrt{3} \approx 1.7321$ and 
$C = 1$ were used to generate data. The result of the training is shown in Figure \ref{2d_training}. 
The learned parameters are $\alpha = 1.72888$ and $C = 0.99977$, within $1\%$ of the true values, with hyperparameters 
$\sigma = 0.0213$, $\theta_1 = 1.581$, and $\theta_2 = 1.586$. 

These examples demonstrate the feasiblity of implementing the fractional kernels as described in the previous sections and the accuracy of discovered parameters, even with noise or small data. Moreover, there is no theoretical difficulty in increasing the dimension of the problem. However, in additional to longer runtime for the computation of formulas \eqref{kernel_theorem_equations}, the user should expect significantly more data to be required for accurate parameter estimation. For example, performing the same two-dimensional example above, with only 20 data points for each of $u$ and $f$, results in learned parameters $\alpha = 1.76516$ and $C = 0.81762$, the $C$ parameter exhibiting an error of roughly 20\%. In the analogous one-dimensional example (Fig. \ref{ex1a}), a $1\%$ error is obtained for less than half this number of data points.} 

{In concluding this section, we point that since the estimation of the equation parameters is based on accurate Gaussian process regression through the training data, there are various well-known hazards to avoid. In addition to obvious issues such as unresolved data and data that is too noisy, too much data in featureless ``flat'' regions carries the risk of overfitting, and should be avoided. } 

\begin{figure}[h]
\centerline{\includegraphics[width=4.5in]{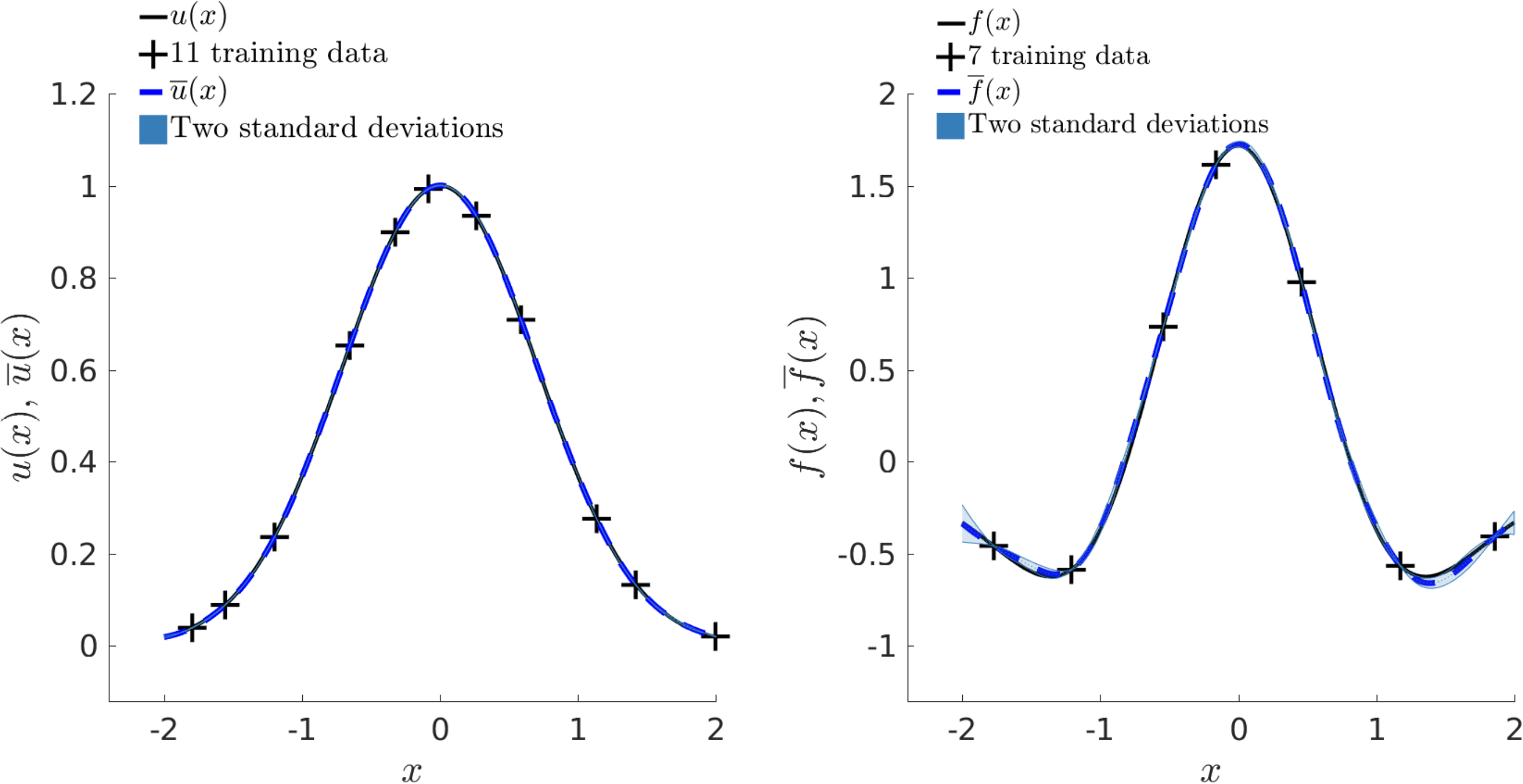}}
\vspace{-1ex}
\caption{\small Result of training the GP in the one-dimensional example on 7 noise-free data points for $u$ and 11 data points for $f$ on $[-2,2]$. The data can be located at arbitrary positions. The trained parameters are 
are within 1\% of the true values. \emph{Optimization wall time: 7 minutes. Roughly 1400 function evaluations.}
\textbf{\emph{All reported wall times in this article are obtained using four threads (default MATLAB vectorization) on an Intel i7-6700k at stock frequency settings}.}
}
\label{ex1a}
\end{figure}

\begin{figure}[h]
\centerline{\includegraphics[width=4.5in]{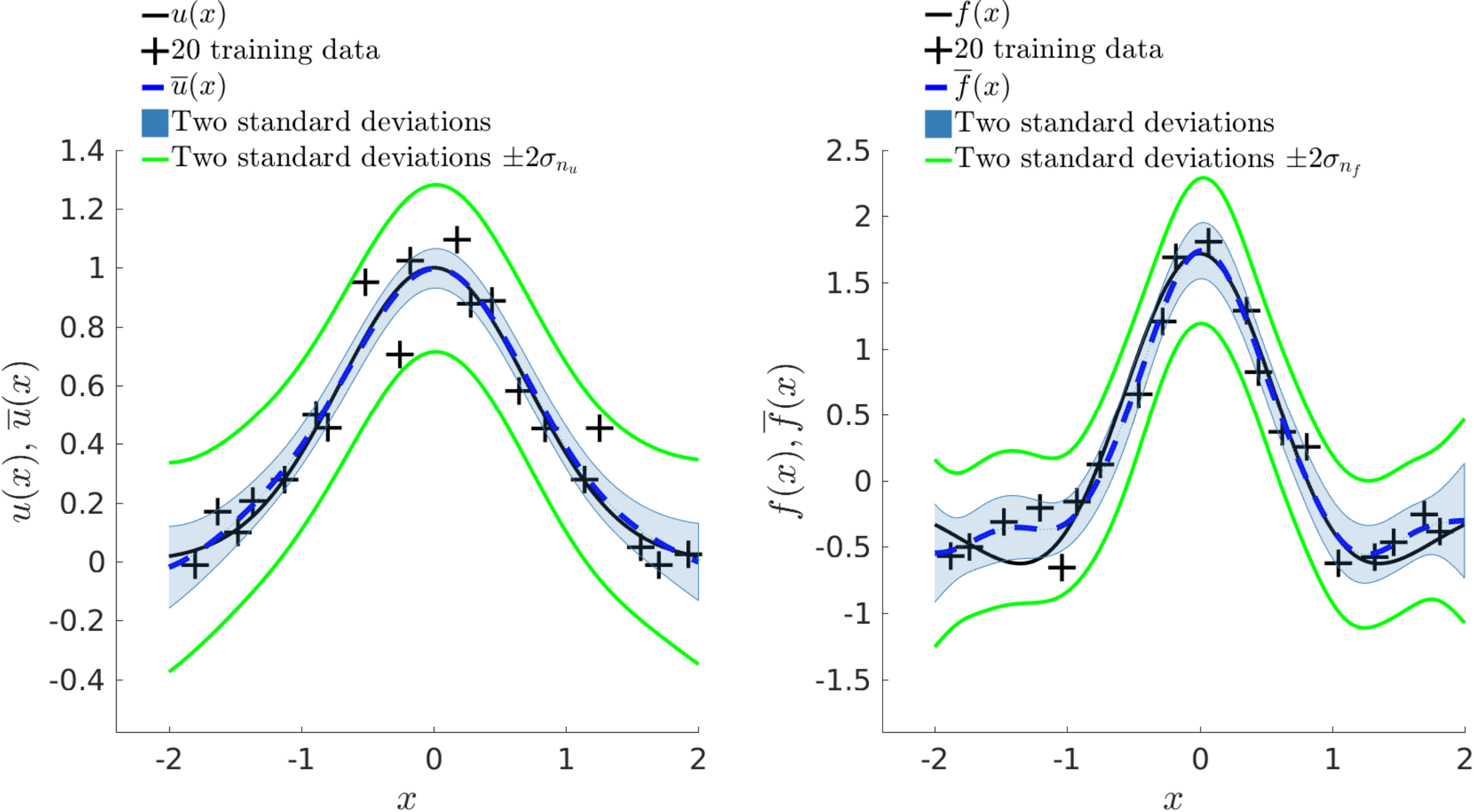}}
\vspace{-1ex}
\caption{\small Result of training the GP on 20 noisy data points for each of $u$ and $f$ on $[-2,2]$. The trained parameters are within 7\% percent of the true values. For this example, we have also plotted in green two standard deviations of the GP, \emph{plus} two times the learned noise parameter $\sigma_{n_{u}} / \sigma_{n_{f}}$. \emph{Optimization wall time: 38 seconds. Roughly 100 function evaluations.}}
\label{ex1b}
\end{figure}

\FloatBarrier

\newpage

\begin{figure}[h]
\center{
\begin{subfigure}{0.48\textwidth}
\centering
\centerline{\includegraphics[width=5.5in]{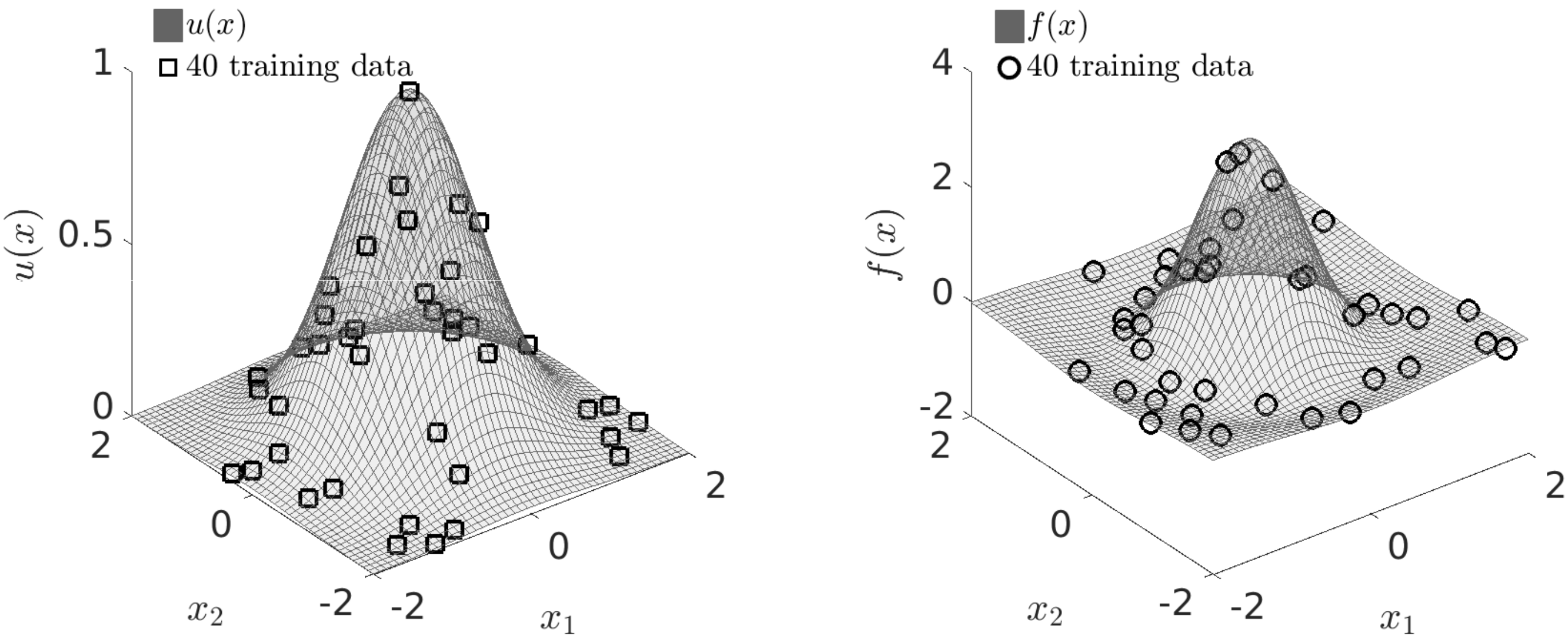}}
%\caption{(40 data points) Surface plots of the exact $u$ (left) and $f$ (right), which solve the equation with parameters $C = 1$ and $\alpha = \sqrt{2}$, that are used to generate data. 40 data points on each of $u$ (squares) and $f$ (circles) are shown on these surface plots.}
\end{subfigure}

\vspace{0.15in}

\begin{subfigure}{0.48\textwidth}
\centering
\centerline{\includegraphics[width=5.5in]{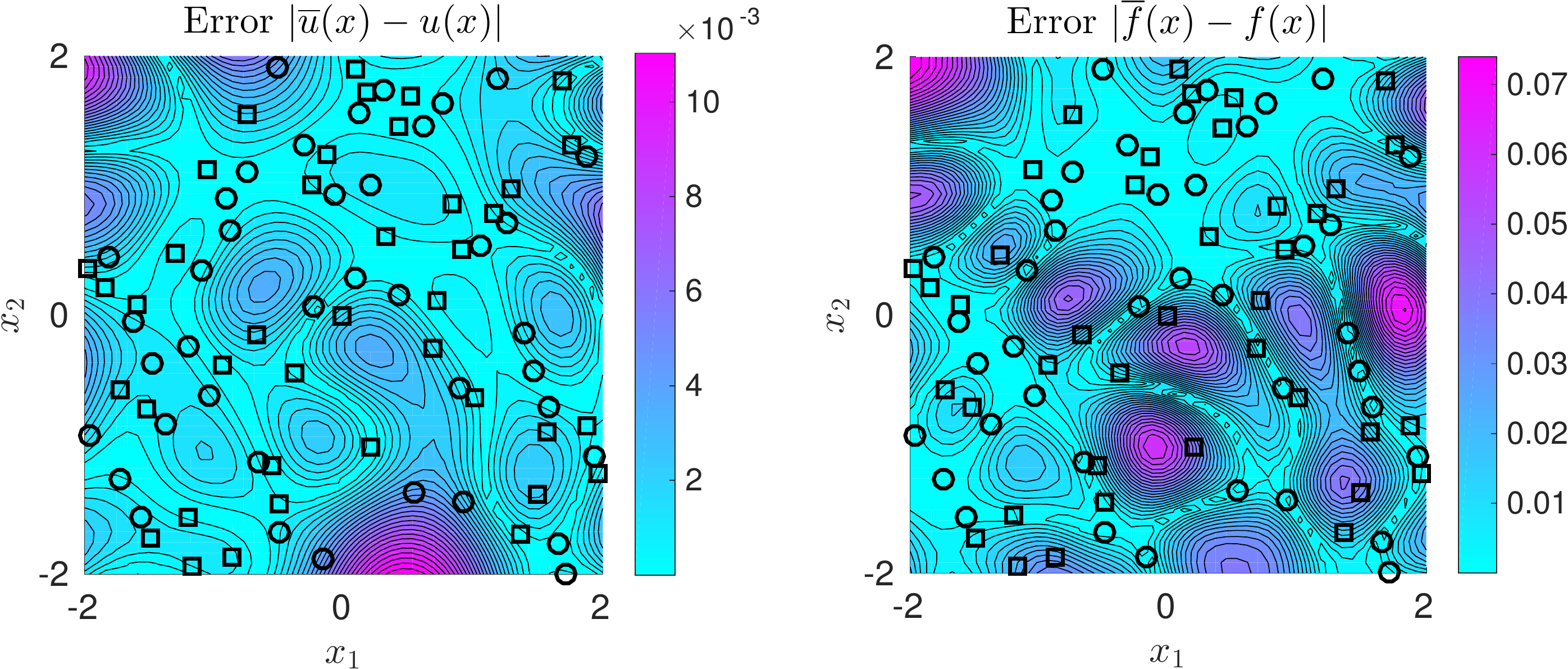}}
%\caption{(40 data points) Error between mean of the trained GP for $u$ (left) and $f$ (right) and the exact $u,f$ used to generate data.}
\end{subfigure}

\vspace{0.15in}

\begin{subfigure}{0.48\textwidth}
\centering
\centerline{\includegraphics[width=5.5in]{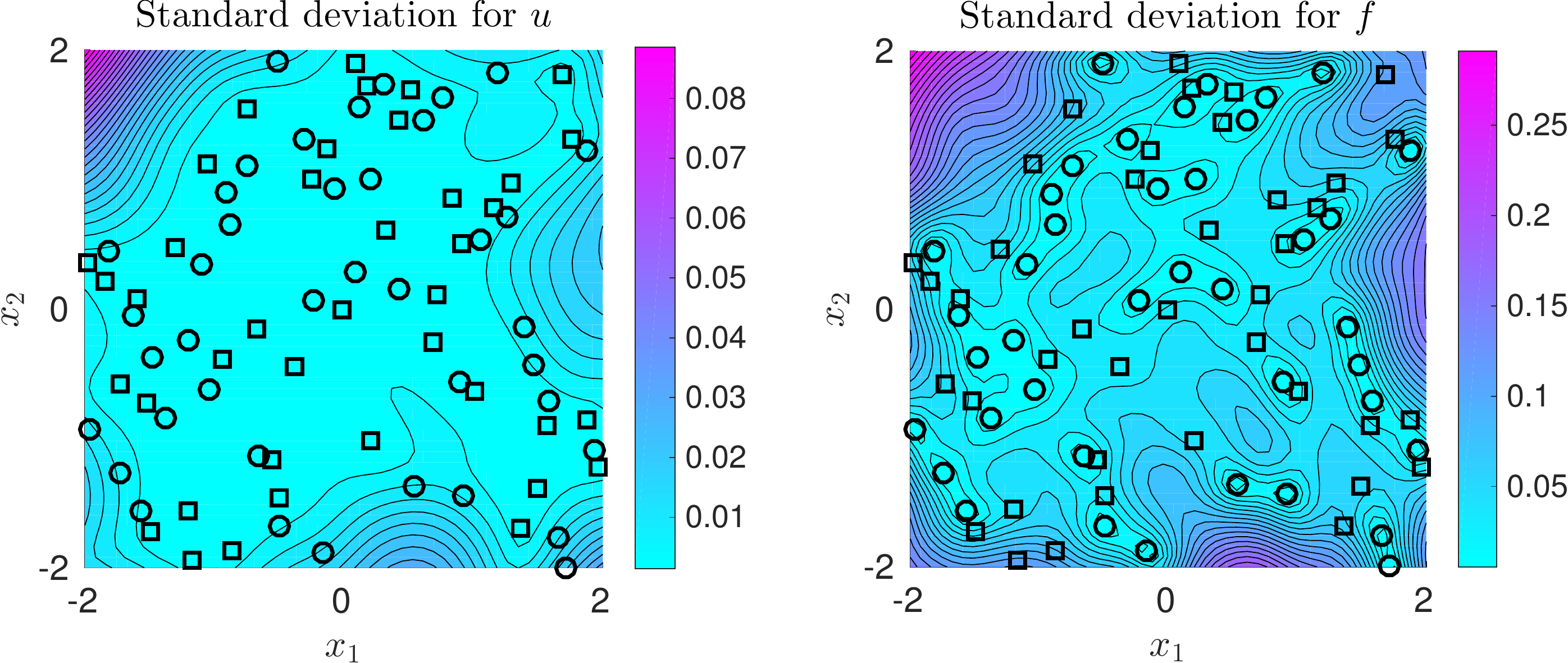}}
%\caption{(40 data points) Standard deviation of the GP for $u$ and $f$.}
\end{subfigure}
\caption{\label{2d_training} Result of training in the two-dimensional example. 
\emph{top:} Distribution of 40 data points on each of $u$, left, and $f$, right. 
\emph{middle:} Error between mean of the trained GP for $u$, left, and $f$, right, and the exact $u,f$ used to generate data. \emph{bottom}: Standard deviation of the GP for $u$, left, and $f$, right. Note that the positions of data points for $u$ are illustrated by squares, while data points for $f$ are illustrated by circles. \emph{Optimization wall time: 74 minutes. Roughly 1400 function evaluations.}}
}
\end{figure}

\newpage

\FloatBarrier

\section{Discovering and Interpolating Integer Order Models}\label{interpolation}
As discussed in the introduction, fractional-order differential operators interpolate between classical differential operators of integer order, reducing the task of choosing a ``dictionary'' of operators of various orders and the assumptions that this entails. The user controls the parsimony directly, by choosing the number of fractional terms in the candidate equation. For example, the model in Section \ref{examples} was constrained to be of parsimony one, since it includes a single differential operator of fractional order. This raises several questions, such as: Can the method be used to discover integer-order operators when they drive the true dynamics? What can be expected if the user-selected parsimony is lower than the parsimony of the ``true'' model driving the data? Can lower-parsimony models still be used to model dynamics?
To explore these questions, we consider the parametrized model
\begin{equation}
\label{fractional_interpolation}
\frac{\partial u}{\partial t} - C \left( ^{RL}_{-\infty} D_{{x}}^{\alpha} \right) u = 0,
\end{equation}
for $u(t,x), t \in \mathbb{R}^+, x \in \mathbb{R}$. 
where the left-sided Riemann-Louiville derivative $^{RL}_{-\infty} D_{{x}}^{\alpha} u$ was defined in \eqref{derivatives_defs}. Note that
\begin{equation}
\alpha = 1 \implies ^{RL}_{-\infty} D_{{x}}^{\alpha} u = -\partial u/\partial x, \quad
\alpha = 2 \implies ^{RL}_{-\infty} D_{{x}}^{\alpha} u = \partial^2 u/\partial x^2.
\end{equation}
For $\alpha = 1, C = 1$, equation \eqref{fractional_interpolation} reduces to the advection equation (with speed $C=1$)
\begin{equation}
\label{integer_transport}
\frac{\partial u}{\partial t} + \frac{\partial u}{\partial x}  = 0,
\end{equation}
while for $\alpha = 2, C = 1$, it reduces to the diffusion equation (with diffusion coefficient $k=1$)
\begin{equation}
\label{integer_heat}
\frac{\partial u}{\partial t} - \frac{\partial^2 u}{\partial x^2} = 0
\end{equation}

We perform four experiments. 
Importantly, we learn these equations using the time-stepping methodology, optimizing \eqref{NLML}, where the kernel blocks are given by \eqref{time_dependent_kernels} and \eqref{time_dependent_kernels_data}. We take $k = M_{\frac{19}{2}}$ (effectively a squared-exponential kernel) and again use \eqref{kernel_theorem_equations} with generalized Gauss-Laguerre quadrature to evaluate the action of the fractional derivatives on $k$.
In all of experiments, we generate data that satisfies the
initial condition $u_0 = \sin(x)$. We choose $\Delta t = 0.1$ and $n =3$; thus, $u^n = u(t = 0.3, x), u^{n-1} = u(t = 0.2, x)$. The GP is trained on 30 data points for each of these time slices. The experiments are
\begin{enumerate}
\item
Data generated from $u(t,x) = \sin(x-t)$, the solution to the advection equation \eqref{integer_transport}. We learn the order $\alpha$ and coefficient $C$ in \eqref{fractional_interpolation}; the exact $\alpha = 1$. 
\item
Data generated from $u(t,x) = e^{-t} \sin(x)$, the solution to the heat equation \eqref{integer_heat}. We learn the order $\alpha$ and coefficient $C$ in \eqref{fractional_interpolation}; the exact $\alpha = 2$.
\item
Data generated from $u(t,x) = e^{-t} \sin(x-t)$, the solution to the integer order advection-diffusion equation 
\begin{equation}
\label{advection_diffusion}
\frac{\partial u}{\partial t} + \frac{\partial u}{\partial x} - \frac{\partial^2 u}{\partial x^2} = 0.
\end{equation}
We learn the order $\alpha$ and coefficient $C$ in \eqref{fractional_interpolation}. Note that this archetype is limited to only one space-differential operator; we will see that the algorithm will select best order $1 < \alpha < 2$ to capture both the advection and diffusion in the data. 
\item
The same advection-diffusion data as in experiment 3, but with the two term, four parameter candidate equation
\begin{equation}
\frac{\partial u}{\partial t} - {C_1} \left( ^{RL}_{-\infty} D_{{x}}^{\alpha_1}  \right) u 
 - {C_2} \left( ^{RL}_{-\infty} D_{{x}}^{\alpha_2} \right) u = 0
\end{equation}
\end{enumerate}
 We note that all of these experiments call for only a \emph{single} fractional-order dictionary term; even
Experiment 4 uses two copies of the same archetype. Experiments 1-3 use initial parameters $\alpha = 0.5$ and $C = 1.25$, and 
Experiment 4 uses $\alpha_1 = 0.5$, $\alpha_2 = 1.5, C_1 = 1.25, C_2 = 0.75$.

In Experiments 1 and 2, we note that fractional-order parameters are discovered close to (within 5\% of) the true integer-order parameters. In this sense, the true dynamics can be considered recovered. The numerical difference from the true parameters despite the high number of data points (30 per slice) is likely due to approximation error from the backwards Euler approximation \eqref{eq:BackwardEuler}, and may be resolved by using higher-order differentiation with more time slices \cite{doi:10.1137/17M1120762}, as simply taking $\Delta t$ to be extremely small may cause the optimization to be dominated by numerical error in computing the kernels. 

\begin{figure}[h!]
\centerline{\includegraphics[width=5in]{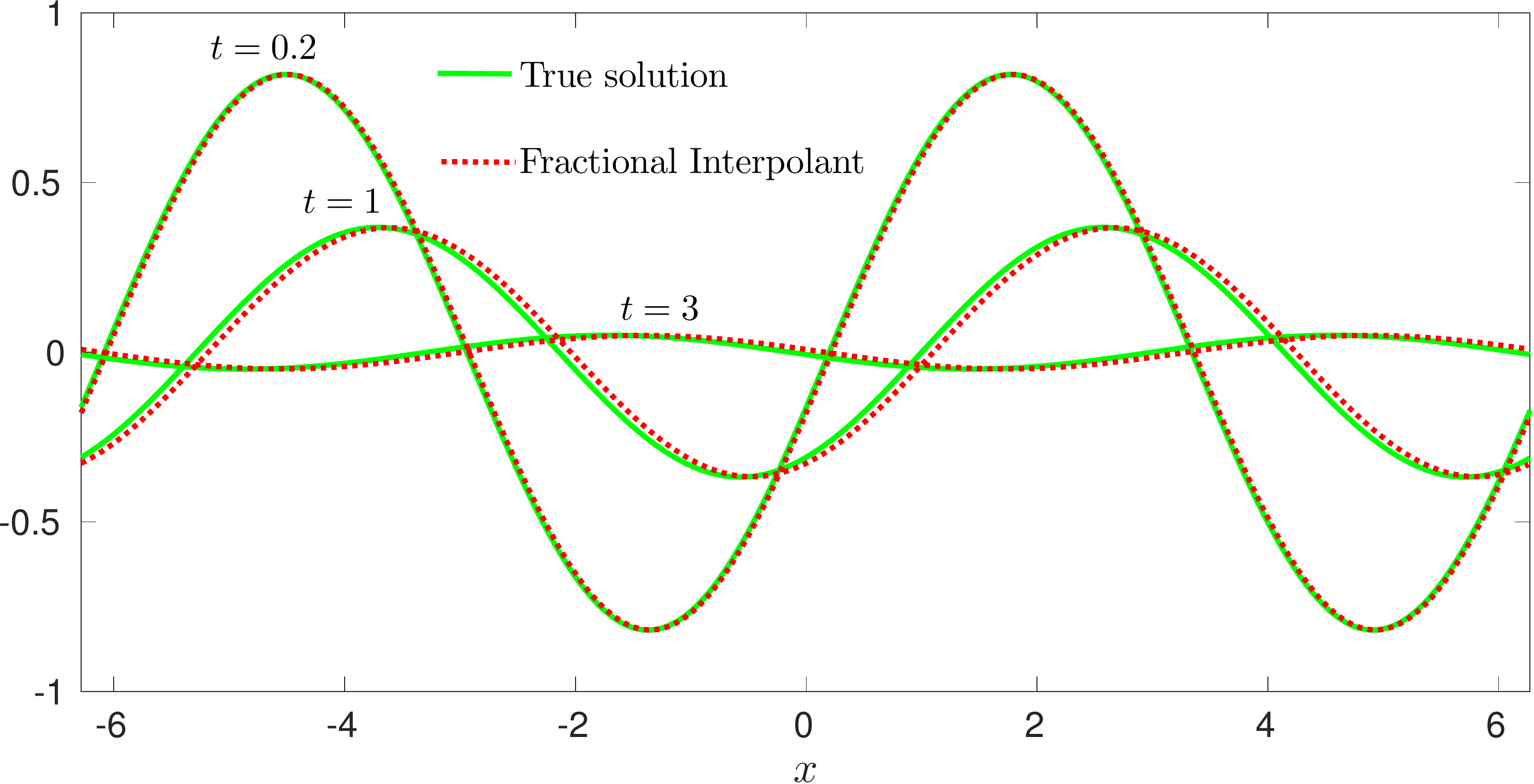}}
\caption{\small Comparison of the true advection-diffusion dynamics with the fractional-order dynamics learned in Experiment 3. Near the time ($t = 0.2$) when the equation was learned, the fractional-dynamics are a good approximation, although for later time $t$ the dynamics are increasingly out-of-phase.}
\label{interpolation_comparison}
\end{figure}

Experiment 3 shows what occurs when the user-defined parsimony is less than the true dynamics used to generate data. The optimizer still converges, and to a sensible answer -- the result can be interpreted as an interpolation of the two integer-order operators in the true dynamics. Moreover, as shown in Figure \ref{interpolation_comparison}, near the time of the data used to train the model, and even much later, the fractional dynamics are a good approximation to the true dynamics, while being simpler in the sense of being driven by only one spatial derivative. This leads to potential applications of interpolating complex systems using lower-parsimony fractional models.  

In Experiment 4, the parsimony was increased with the inclusion of an additional indepedent copy of the fractional archetype. The advection-diffusion equation is recovered with parameters within $5\%$ of the true values. Thus, with a single fractional archetype and user-controlled parsimony, it is possible to discover advection, diffusion, advection-diffusion, as well as a single-term fractional interpolation of advection-diffusion.

\setlength{\tabcolsep}{0.2em} % for the horizontal padding
{\renewcommand{\arraystretch}{1.5}% for the vertical padding
\begin{table}[h!]
\small
\centering
\caption{\small Results of the four experiments. \emph{Wall times: roughly 11/14/13/21  minutes.} }
\label{four_experiments}
\begin{tabular}{|p{0.05\textwidth}|p{0.25\textwidth}|p{0.45\textwidth}|p{0.15\textwidth}|}
\hline
Exp. & Data                & Candidate   & Parameters Learned \\ \hline
1   & Advection: \linebreak
$
\frac{\partial u}{\partial t}  + \frac{\partial u}{\partial x} = 0 
$ 
& 
$ \frac{\partial u}{\partial t} - C \left( ^{RL}_{-\infty} D_{{x}}^{\alpha} u \right) = 0 $
 & $C = 1.01 \newline \alpha = 0.97$ \\ \hline
2   & Diffusion: \linebreak
$
\frac{\partial u}{\partial t} - \frac{\partial^2 u}{\partial x^2} = 0
$
& 
$ \frac{\partial u}{\partial t} - C \left( ^{RL}_{-\infty} D_{{x}}^{\alpha} u \right) = 0 $
& $C = 1.05 \newline \alpha = 2.00$ \\ \hline
3   & Advection-Diffusion: $\frac{\partial u}{\partial t} + \frac{\partial u}{\partial x} - \frac{\partial^2 u}{\partial x^2} = 0 $ 
&
$\frac{\partial u}{\partial t}  - C \left( ^{RL}_{-\infty} D_{{x}}^{\alpha} \right) u = 0$
& $C = 1.49 \newline \alpha = 1.47$ \\ \hline
4   & Advection-Diffusion: $\frac{\partial u}{\partial t} + \frac{\partial u}{\partial x} - \frac{\partial^2 u}{\partial x^2} = 0 $ 
&
$\frac{\partial u}{\partial t} - {C_1} \left( ^{RL}_{-\infty} D_{{x}}^{\alpha_1}  \right) u 
 - {C_2} \left( ^{RL}_{-\infty} D_{{x}}^{\alpha_2} \right) u = 0$
& $C_1 = 1.05 \newline \alpha_1 = 0.98$ \newline $ C_2 = 1.03 \newline \alpha_2 = 1.96$ \\ \hline
\end{tabular}
\end{table}
}

%\iffalse{

\section{Learning Fractional Diffusion from $\alpha$-stable Time Series}\label{variable_order_matern}
We now consider an example that will set up our application to financial time series data and explore the effect of the Mat\'ern parameter $\nu$. The example will involve identification of $\alpha$-stable L\'evy processes parameters from time series data. Equivalently, the same data is used to identify a fractional-order diffusion equation governing the transition density of the process, of the form
\begin{equation}
\label{diffusion}
\frac{\partial u}{\partial t} = 
\frac{\gamma^\alpha}{|\cos(\pi \alpha / 2)|}
\left[ p (^{RL}_{-\infty} D_{{x}}^{\alpha} u) + (1-p) (^{RL}_{{x}}
D^{\alpha}_{\infty}u) \right], \quad t > 0.
\end{equation}
where 
\begin{equation}
0 < \gamma, \quad 0 < p < 1, \quad 0 < \alpha < 2.
\end{equation}
The exact solution to this equation (technically, with intial condition a point distribution at zero) is in fact the $\alpha$-stable probability density 
\begin{equation}
\label{stable_density}
S_\alpha(2p-1, \gamma t^{1/\alpha}, 0)
\end{equation}
with stability parameter $\alpha$, skewness parameter $2p-1$, 
scale parameter $\gamma t^{1/\alpha}$,  and position parameter $0$.
See Proposition 5.8 in Meerschaert and Sikorskii \cite{meerschaert_sikorskii}.

\begin{figure}[h!]
\centerline{\includegraphics[width=5.5in]{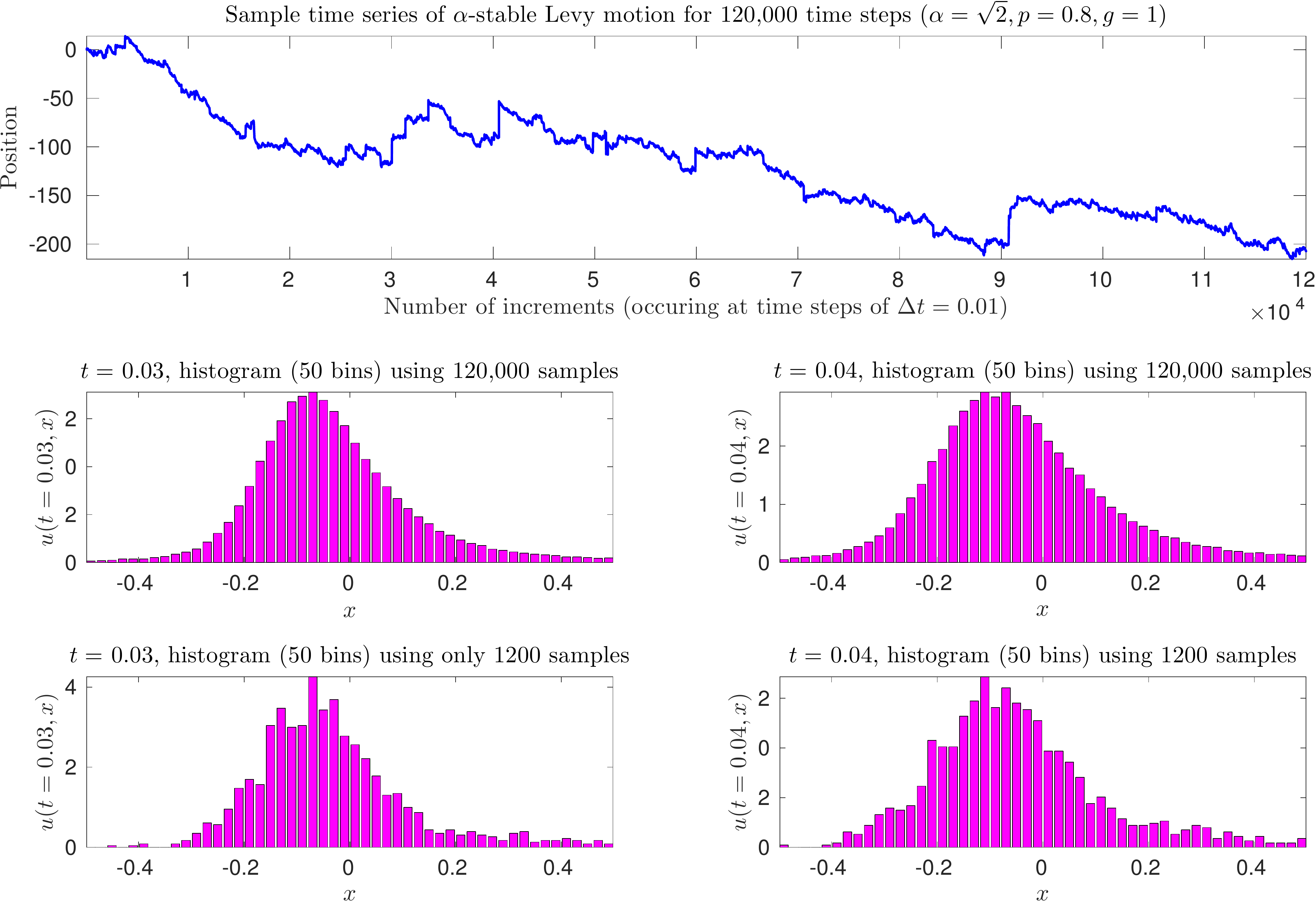}}
\caption{
\small
Illustration of how data is produced to discover a fractional-order diffusion equation using time series data. The top shows an example $\alpha$-stable time series with $\alpha = \sqrt{2}, p = 0.8$, $\gamma = 1$ at time increments of $\Delta t = 0.01$. 
To approximate the solution at $t = 0.03$, as on the left, a histogram is made of the spatial increments in three units of time along the time series, and normalized. The longer the time series, the more accurate the empirical density will approximate the true density/solution. The middle row shows histograms made with a time series of 120,000 increments, while the bottom shows histograms made with a time series of 1,200 increments. 
\label{empirical}
}
\end{figure}

Under the ansatz that the increments are drawn from a transition density, a time series $X_i$, $i = 1, ..., i_{\text{max}}$ can be used to recover the transition densities in the following way. Suppose that the increments occur in units of time $\Delta t$.
To approximate the transition density $\rho_{n \Delta t}$ at times $n \Delta t$, $n \in \mathbb{N}$, first the collection of increments
\begin{equation}
\{ X_{(i+n)\Delta t} - X_{i\Delta t }\}_{i = 1, 2, ..., i_{\text{max}} - k}.
\end{equation}
is assembled. A histogram of such increments is created and normalized; the result is an empirical probability distribution, and as the the number of samples increases, the empirical distribution converges to the probability density function \cite{vaart_1998}. 
Therefore, if $X_t$ is a time series in which, at each time increment $\Delta t$, a space increment is drawn from the 
$\alpha$-stable density with parameters skewness $2p-1$, scale $\gamma (\Delta t)^{1/\alpha}$, and position $0$, then these empirical histograms $\rho_{n \Delta t}$ will approximate the same density
${S_\alpha(2p-1, \gamma (n \Delta t)^{1/\alpha}, 0)}$. In other words,  $\rho_{n \Delta t}$ will approximate the time-slices $u(t = n\Delta t,x)$ of 
the solution to equation \ref{diffusion}. This setup is shown in Figure \ref{empirical}, which illustrates empirical distributions using 1,200 and 120,000 samples. 

\begin{figure}
 \centering
  \subcaptionbox{}{\includegraphics[width=2.5in]{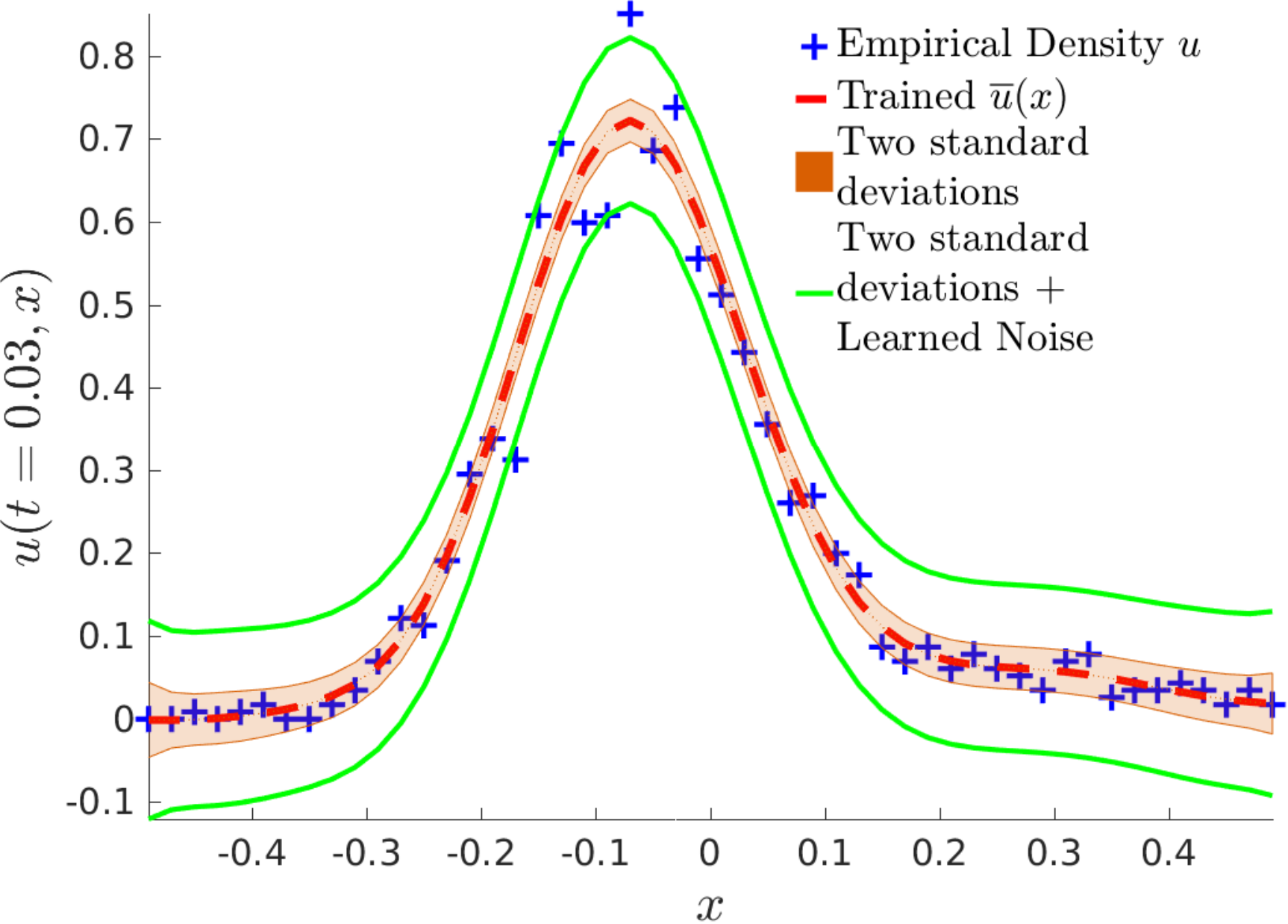}}\hspace{0em}%
  \subcaptionbox{}{\includegraphics[width=2.5in]{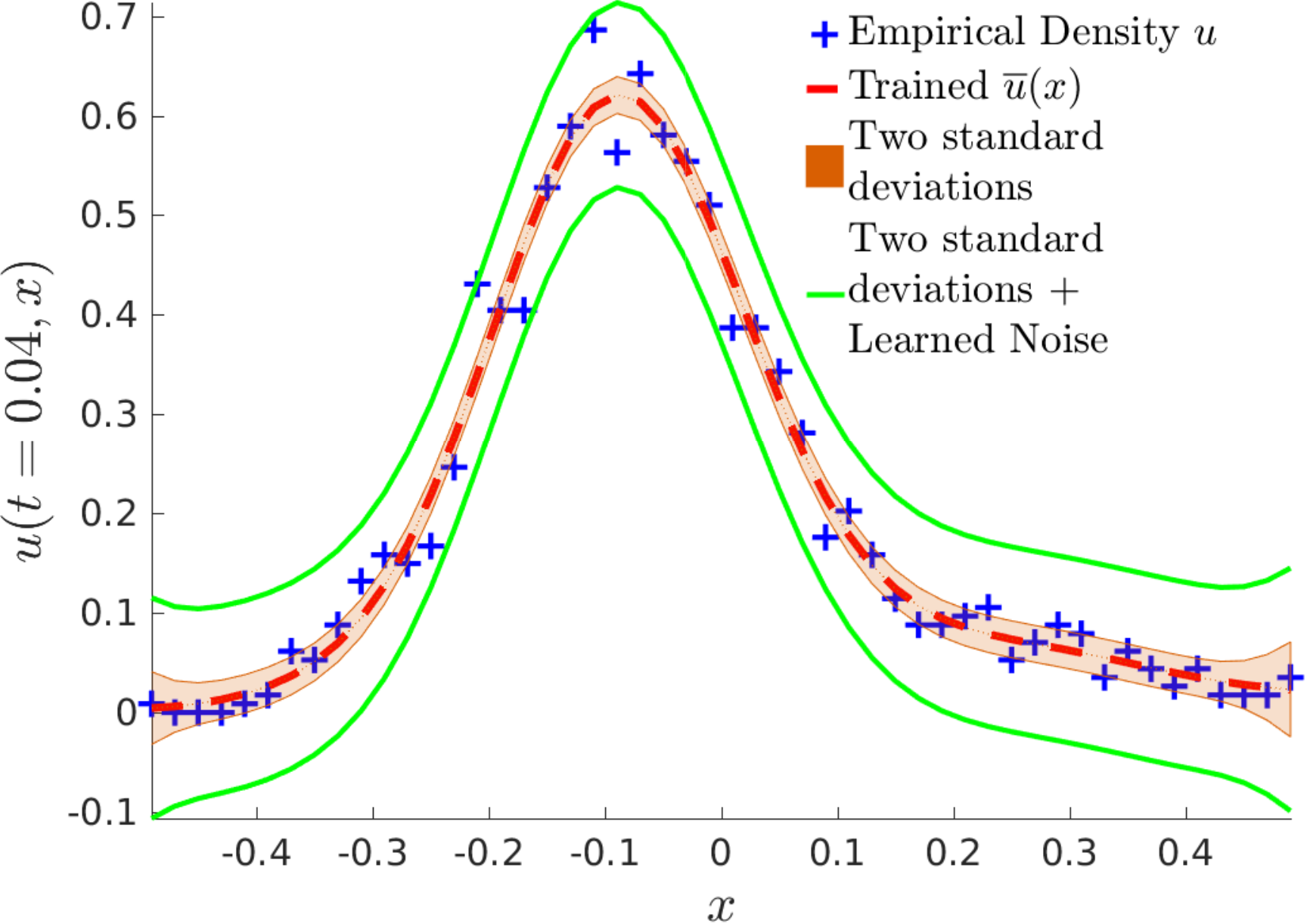}}
\caption{\small \label{synthetic_time_series_training} Result of training the GP on the synthetic 1200-step $\alpha$-stable time series data. The blue crosses are the normalized histogram/empirical distribution function data. The red curve/orange bars show the mean/two standard deviations of the trained GP, while the green curve shows two standard deviations $\pm$ the learned noise $2\sigma_n$.}
\end{figure}

In preparation for the next section, we will use the much noiser data generated from an $\alpha$-stable time series of 1,200 steps and parameters $\alpha = \sqrt{2}, p = 0.8$, $\gamma = 1$ and $\Delta t = 0.01$, at times $t = 0.03$ and $t = 0.04$. This is the data shown in the bottom panel of Figure \ref{empirical}. Because we are only seeking an equation that can be associated with an $\alpha$-stable process, we enforce $0\le p \le 1$ and $0 < \alpha \le 2$ by placing these variables in sigmoids:
\begin{equation}
p = \frac{1}{1+\exp[{-\tilde{p}}]}, \quad
\alpha = \frac{2}{1+\exp[{-\tilde{\alpha}}]}.
\end{equation}
The optimization is then over $\tilde{p}$ and $\tilde{\alpha}$.
The parameters are initialized as $\alpha = 1.8$, $p = 0.5$, and $\gamma = 1$. Because the data appears quite rough, we have also allowed for variable order $\nu$, optimizing it as an extra hyperparameter, initializing with $\nu = 9/2$. 

The result of the training is shown in \ref{synthetic_time_series_training}. The trained parameters are $\alpha = 1.38415, 
p = 0.87519, \gamma = 0.89166, \sigma = 0.05576, \theta = 0.16287$, and $\nu = 14.82891$. First, we note that the equation parameters are roughly within $10$\% of the values used to generate the data, despite the low number of samples and roughness of the histogram. Next, the learned value of $\nu$, which is much higher than the starting value of $4.5$, shows that the Gaussian process was able to learn the noise parameter $\sigma$ and write off the roughness of the histogram as not intrinsic to the data, so that a low $\nu$ value was not required. As higher Mat\'ern kernel $M_\nu$ are for practical purposes the same as the squared-exponential kernel and with each other \cite{rasmussen_williams}, we conclude that for such data it is not necessary to utilize a variable-order Mat\'ern kernel, as it prolongs the optimization needlessly.  Indeed, performing the same simulation with $\nu$ fixed as $9/2$ yields parameters $\alpha = 1.32148, p = 0.80294, \gamma = 0.98273$, which are arguably slightly closer to the true values. However, we do not claim this to be true for all applications; in particular, in cases where enough data is available to resolve local behavior clearly, variable-order $\nu$ optimization may be critical in training the Gaussian process \cite{stein}.

\section{Fractional Diffusion for Relative Stock Performance}\label{finance}
Many types of  L\'evy processes -- such as $\alpha$-stable processes -- are well known and heavily used in financial modeling due to their heavy tails \cite{cont_tankov, Haas2011, BRADLEY200335}. This began with the work of Mandelbrot in 1963 \cite{mandelbrot}, who showed that returns of cotton prices are more accurately modelled by an $\alpha$-stable density with $\alpha = 1.7$ than with a normal density, followed by the work of Fama in 1965 \cite{fama} arriving at a similar conclusion for daily returns of the Dow Jones Industrial Average. More recent investigations include $\alpha$-stable behavior in Mexican financial markets \cite{2012PhyA..391.2990A}, as well as financial modeling by more general  L\'evy processes \cite{DBLP:journals/siamsc/MatacheSW05, doi:10.1137/120881063}. The implications of heavy tailed behavior in financial processes cannot be underestimated in practice; Wilmott \cite{wilmott} gives the following example based on daily data of the S\&P 500 from 1980-2004. While a 20\% fall in the S\&P 500 occured once in this interval of 24 years (the stock market crash of October 19th, 1987) a normal distribution for S\&P 500 returns (based on an average volatility of 16.9\%) would imply such an event would occur only once in every $2\times10^{76}$ years.

A number of methods have been used to determine the parameters of an $\alpha$-distribution from empirical data (see Chapter 7 of \cite{cont_tankov} for a summary, as well as \cite{art_fitting}). A naive approach would be to plot the empirical densities in log-log scale, where the power-law tail would appear linear, and read the stability parameter $\alpha$ from the slope the curve for large argument. This can be misleading because it is not clear for what arguments an $\alpha$-stable distribution is converged to a power law, nor is the answer simple; it depends on the parameter $\alpha$, and the density enters into a transitory power law decay with power $> 2$ before settling into the ``true'' power law decay with $\alpha < 2$ (see \cite{borak}). Much more reliable methods include maximum likelihood estimation \cite{Nolan} and the generalized method of moments \cite{chausse}, which have had good success in practice. 

\begin{figure}[h]
\centerline{\includegraphics[width=5in]{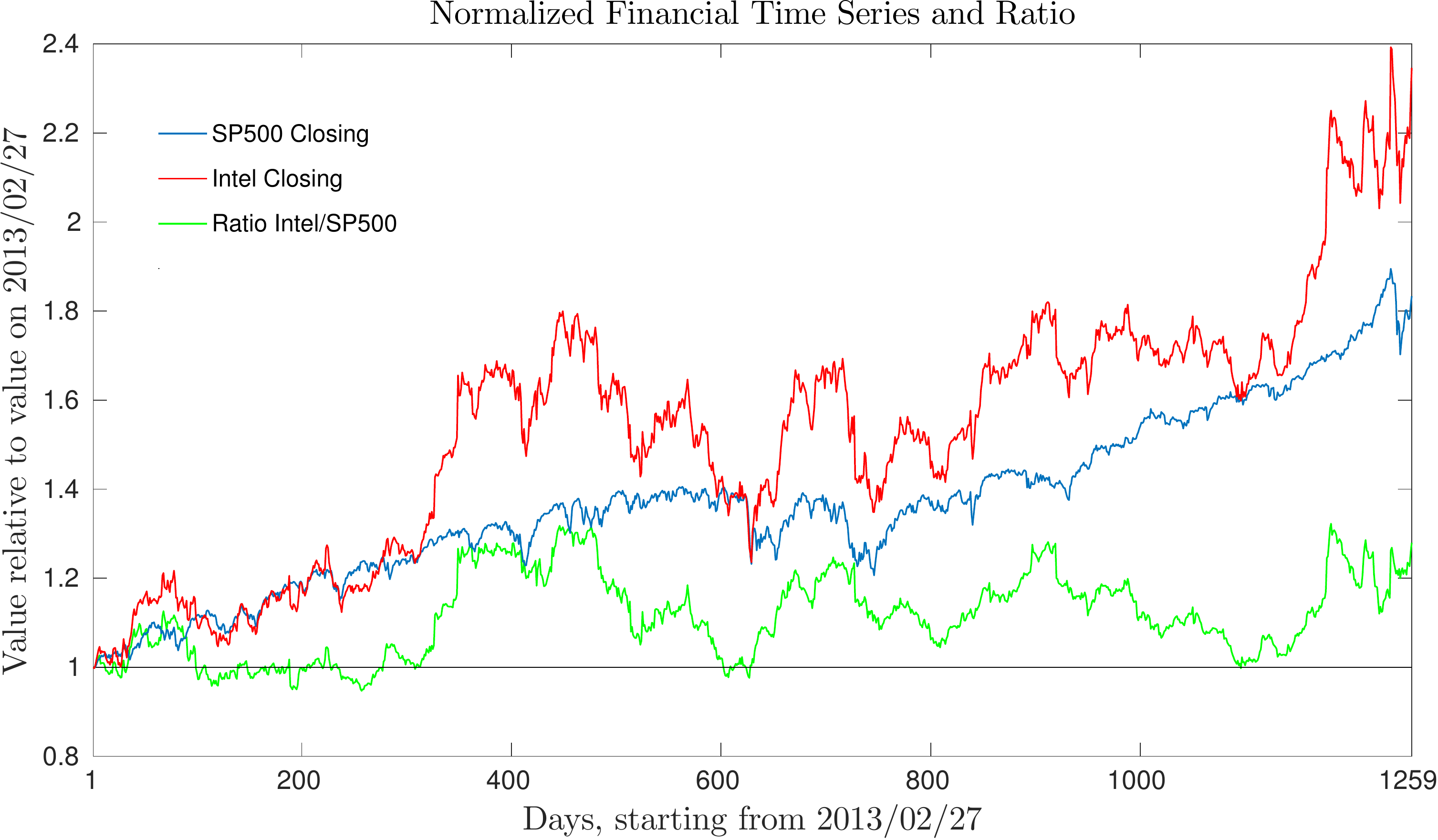}}
\caption{
\small
The financial time series data used to calibrate the model. The blue curve shows the daily closing values of the S\&P 500, normalized to the value on 2013/02/27, while the red curve shows the same for Intel. The ratio (which measures the performance of Intel relative to the S\&P 500 index in the same time period) is shown in green.
\label{fig:three_time_series}
}
\end{figure}

We propose using machine learning of fractional diffusion equations outlined in section \ref{variable_order_matern}, with empirical distributions as data, to calibrate the parameters of the corresponding $\alpha$-stable  L\'evy process. We have demonstrated that such a method can reliably recover the parameters of synthetic data, and can handle very noisy or rough data in  Section \ref{variable_order_matern}. The following simulations use the exact same setup, only with empirical (rather than synthetic) data, to learn the parameters of the equation
\begin{equation}
\frac{\partial u}{\partial t} = 
\frac{\gamma^\alpha}{|\cos(\pi \alpha / 2)|}
\left[ p (^{RL}_{-\infty} D_{{x}}^{\alpha} u) + (1-p) (^{RL}_{{x}}
D^{\alpha}_{\infty}u) \right], \quad t > 0.
\end{equation} 
The raw data we take for illustration is the relative stock performance of Intel vs S\&P 500. We use the daily closing values of each stock in a 5-year period from 2013/02/27 to 2018/02/26. We normalize each stock to the ``initial'' value on 2013/02/27, then take ratio of the two stocks (Intel/SP500) to yield the time series that will be trained on. The time series used in the procedure are illustrated in Figure \ref{fig:three_time_series}. 

\begin{figure}[h]
\centerline{\includegraphics[width=5in]{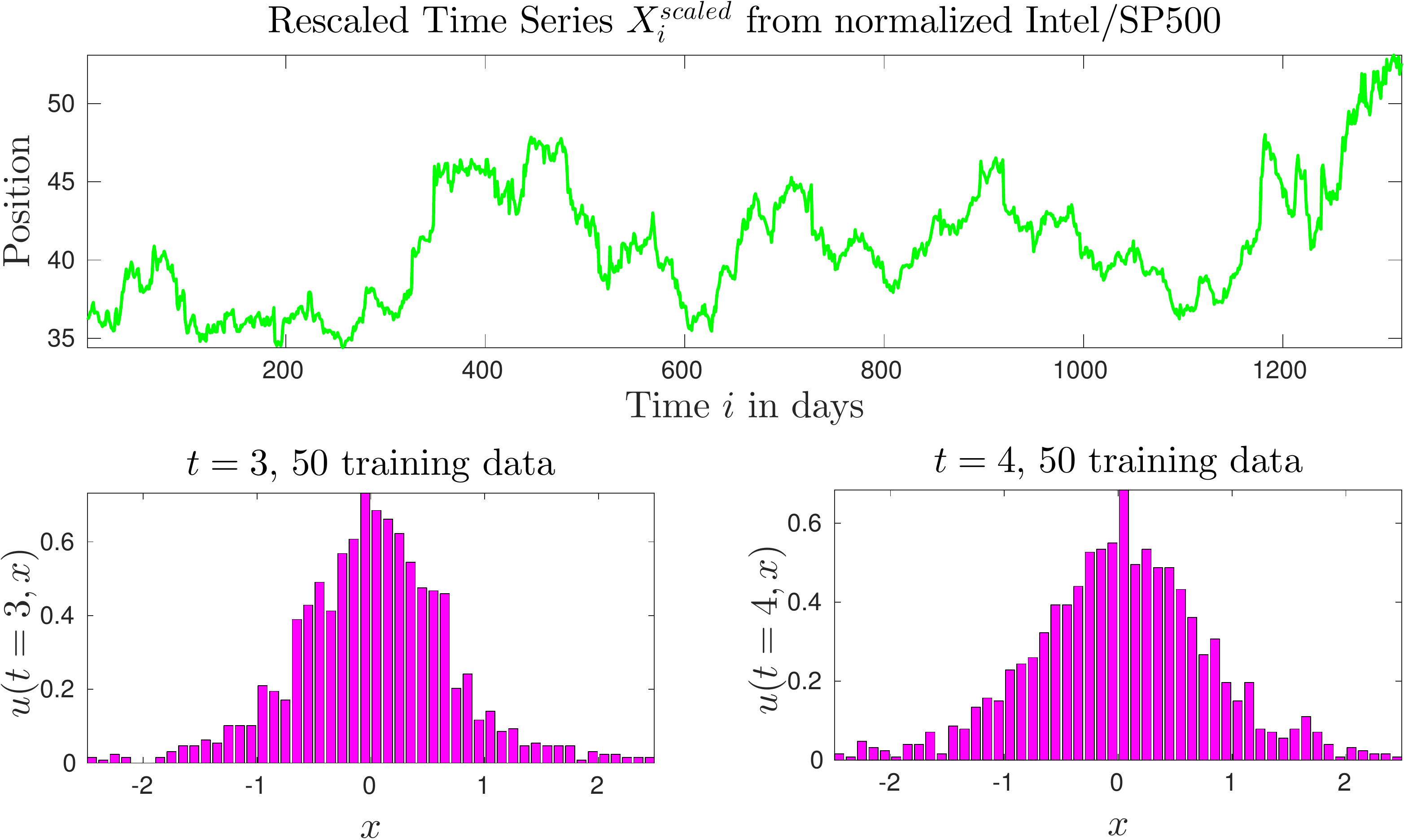}}
\caption{
\small
(top) The rescaled time series given by \eqref{scaled_time_series}. The relative (normalized) Intel/SP500 stock is multiplied by $\sqrt{1259}$ to put it in the same window as standard Brownian motion. (bottom) The empirical histograms to be used as data to discover the fractional-order diffusion equation. 
\label{finance_data}
}
\end{figure}

\begin{figure}
 \centering
  \subcaptionbox{}{\includegraphics[width=2.5in]{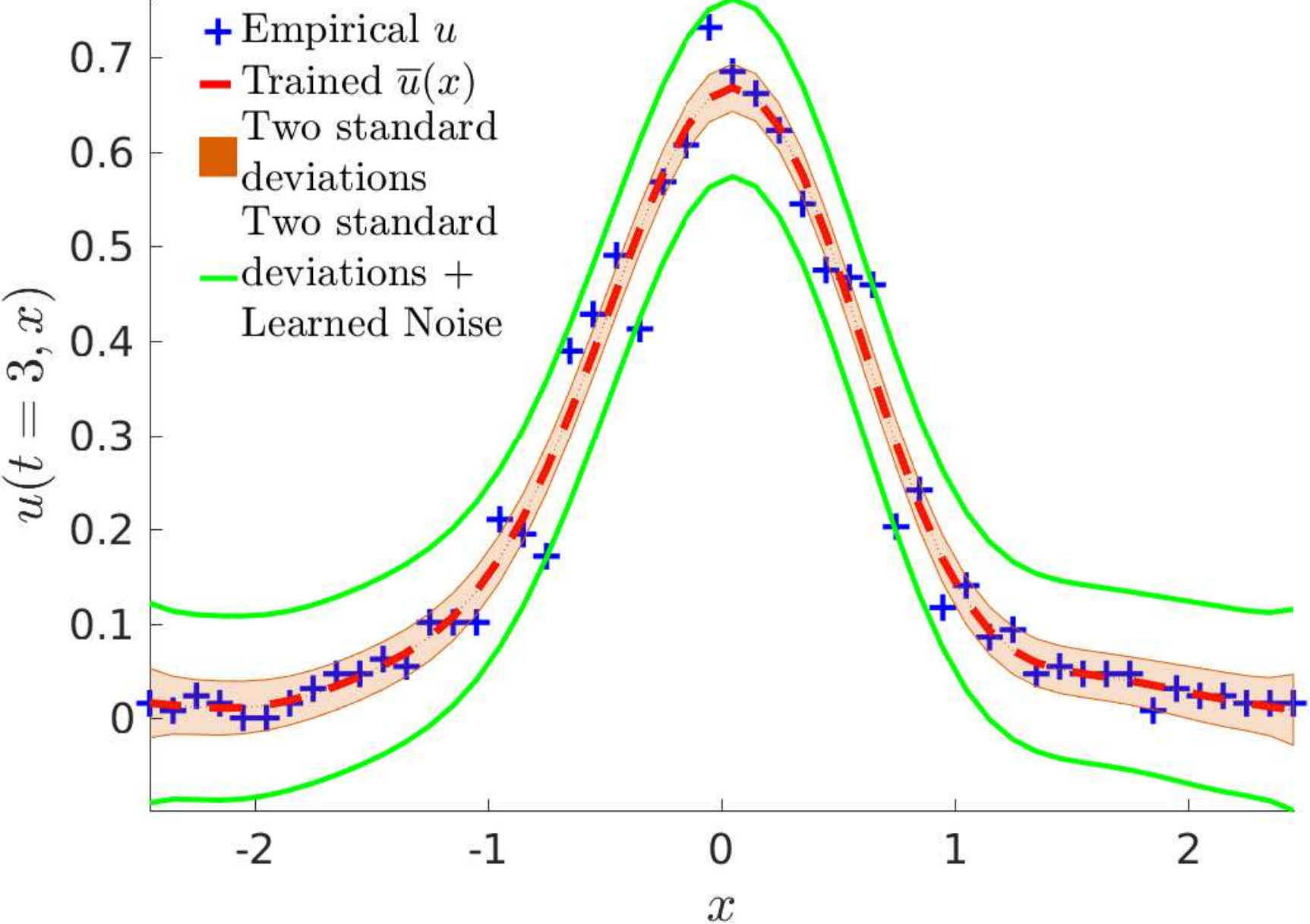}}\hspace{0em}%
  \subcaptionbox{}{\includegraphics[width=2.5in]{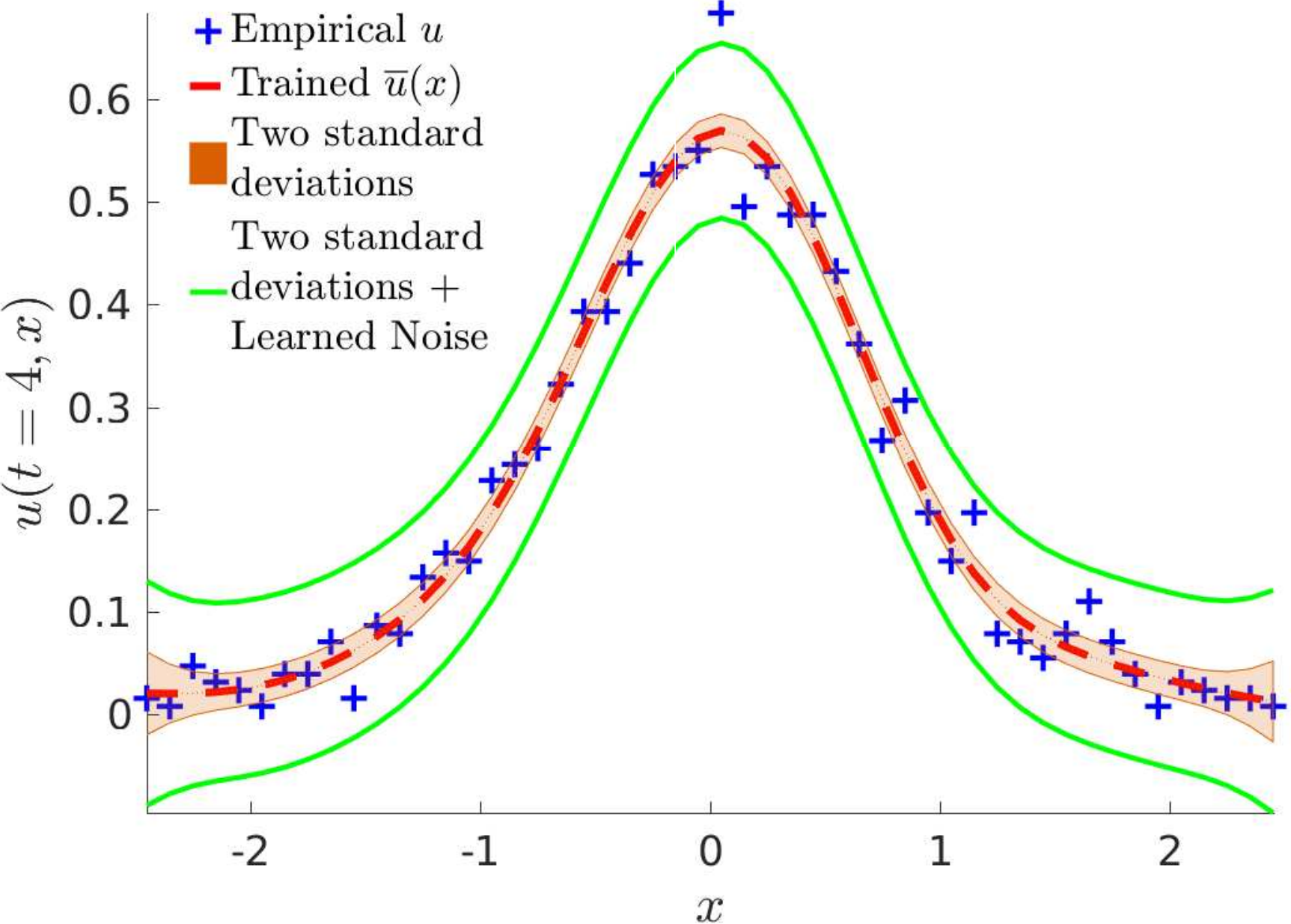}}
\caption{
\small
The result of training the GP using the histogram data in Figure \ref{finance_data}.
The blue crosses are the normalized histogram/empirical distribution function data. The red curve/orange bars show the mean/two standard deviations of the trained GP, while the green curve shows two standard deviations $\pm$ the learned noise $2\sigma_n$.
\label{finance_training}
}
\end{figure}

The natural time intervals here are $\Delta t  = 1$ days. Thus, in \eqref{eq:BackwardEuler} and in the kernels, $\Delta t = 1$, and $t = n$, where $n \in \mathbb{N}$. 
The parameter $\gamma$ can, in principle, capture any space-time scaling relation for the diffusion. However, very small or very large values of $\gamma$ cause optimization instability. Thus, the time series values are rescaled by a constant factor factor $\sqrt{i_{\text{max}}}$ -- the scaling for Brownian motion -- prior to fine tuning with $\gamma$:
\begin{equation}
\label{scaled_time_series}
X_i^{\text{scaled}} = \sqrt{i_{\text{max}}} X_i = \sqrt{1259} X_i.
\end{equation}
This ensures $\gamma = \mathcal{O}(1)$, because it places $X_i$ in roughly the same window as a standard $2$-stable process (Brownian Motion) with the same number of steps. After this preprocessing, a value of $\gamma^{\text{scaled}}$ will be learned for the scaled diffusion. The original diffusion then has true scaling parameter 
\begin{equation}
\gamma = \gamma^{\text{scaled}}/\sqrt{i_{\text{max}}} =   \gamma^{\text{scaled}}/\sqrt{1259}.
\end{equation} 
This is shown in Figure \ref{finance_data}, and the result of the training is shown in Figure \ref{finance_training}. 
Following the discussion in Section \ref{variable_order_matern}, a fixed Matern parameter $\nu = 9/2$ was used. The initial parameters were 
$\alpha = 1.8, p = 0.5, \gamma^{\text{scaled}} = 1, \sigma = 0.1, \theta = 0.2$. The trained parameters were $\alpha = 1.667$, $p = 0.422$,  $\gamma^{\text{scaled}} = 0.235$, and hyperparameters
$\sigma = 0.0617$, $\theta = 1.173$ and learned noise $\sigma_n =  0.0344$. The true $\gamma = 0.0066$. The optimization wall time was 6 minutes and 31 seconds. 

\begin{figure}[h]
\centerline{\includegraphics[width=5in]{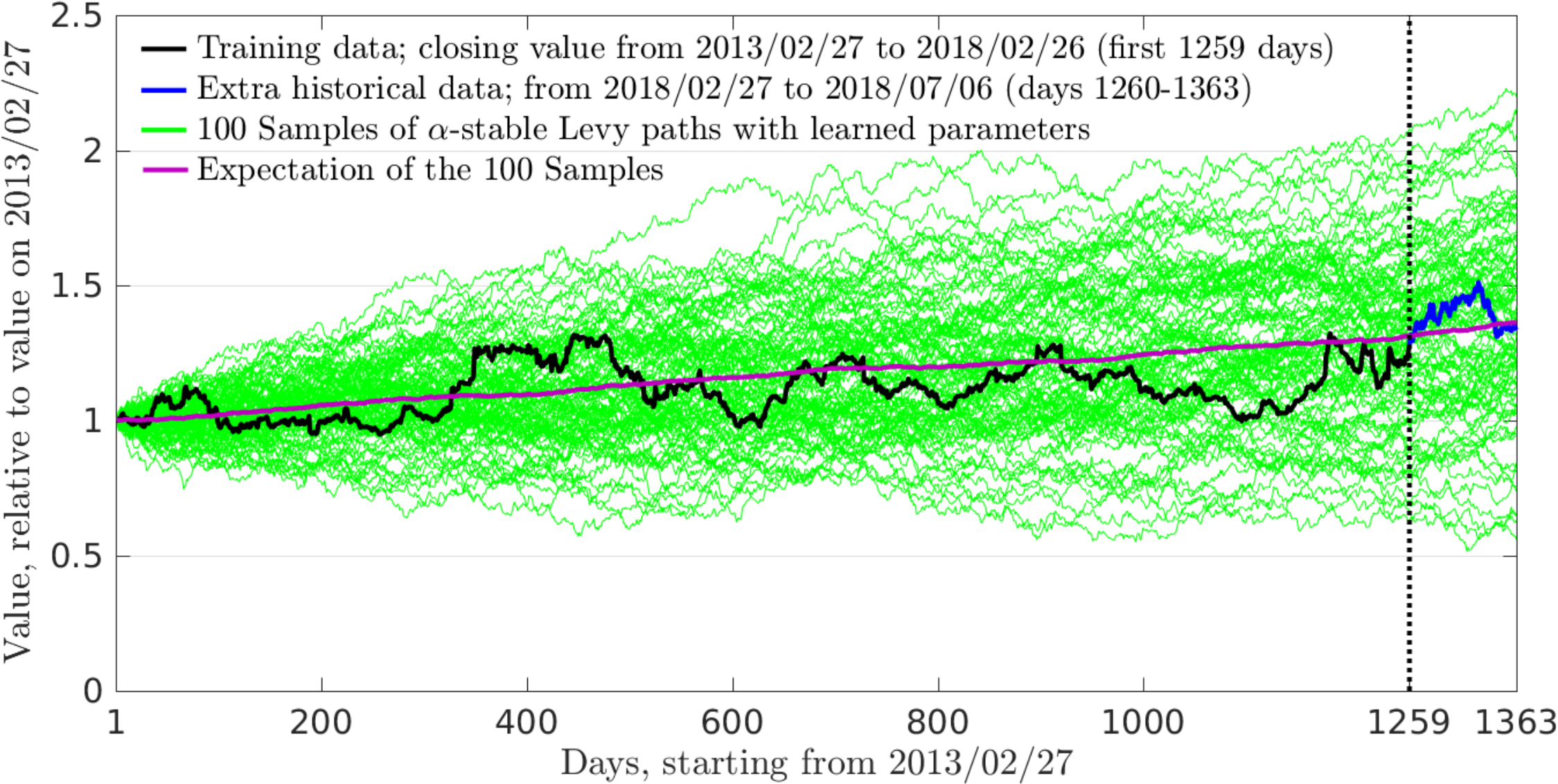}}
\caption{
\label{validation}
\small
Backtesting/validation of the trained $\alpha$-stable parameters. The black curve shows the training data of normalised Intel / normalized S\&P 500 for 1259 days starting Feb. 27, 2013. The blue curve shows extra historical time series data, from Feb. 26, 2018 until July 26, 2018 (not used to train). Each of the 100 green curves is a sample path of the truncated $\alpha$-stable process \eqref{final_distribution}. The dark/filled region of the envelope of samples contains the historical data, providing a fairly sharp estimate of volatility in the first year. The purple curve is the expectation of the samples, which is accurate initially but does not take into account the mean-reverting behavior of the stock over five years.  
}

\vspace{-0.15in}

\end{figure}

The conclusion of this training is that, in this model, the transition density $u(t,x)$ of the original Intel(normalized)/SP500(normalized) process is governed by the fractional diffusion equation
\begin{align}
\begin{split}
\frac{\partial u}{\partial t} &= 
\frac{0.0066^{1.667}}{|\cos(1.667 \pi/ 2)|}
\left[ 0.422 (^{RL}_{-\infty} D_{{x}}^{1.667} u) + 0.578 (^{RL}_{{x}}
D^{1.667}_{\infty}u) \right] \\
& =  0.000113 (^{RL}_{-\infty} D_{{x}}^{1.667} u) + 0.000155 (^{RL}_{{x}}
D^{1.667}_{\infty}u).
\end{split}
\end{align}
Equivalently, the time series may be described as an $\alpha$-stable process,
\begin{equation}
\label{learned_density}
X_{i+n} - X_i \sim S_{1.667}(-0.156,0.0066(n^{1/1.667}),0).
\end{equation}
However, for modelling purposes, it should be kept in mind that the training was performed using data in the interval $[-2.5, 2.5]/\sqrt{1259}$ with $n = 3$ and $n = 4$; increments greater than $2.5/\sqrt{1259}$ in four units of time were excluded. Thus, for consistency, the stable densities should be truncated to this interval as well, and renormalized. If increments in four units of time are drawn from the $1.667$-stable density \eqref{learned_density} with $n = 4$ truncated to $[-2.5, 2.5]/\sqrt{1259}$, increments in one unit of time should be drawn as from the appropriate ($n = 1$) density truncated to 
$[-2.5, 2.5] / (\sqrt{1259} \times 4^{1/1.667}) = [-0.031, 0.031]$ and normalized:
\begin{equation}
\label{final_distribution}
X_{i+1} - X_{i} \sim C_{\text{norm}} \mathbbm{1}_{ [-0.031, 0.031] } S_{1.667} (-0.156,0.0066,0)
\end{equation}
Sampling of this distribution can be performed by sampling $S_{1.667} (-0.156,0.0066,0)$ and rejecting draws greater in magnitude than 
0.031. Backtesting with this model, shown in Figure \ref{validation}, yields good agreement with both the trend and volatility of the historical data.  

Fitting $\alpha$-stable densities to financial data is important for risk management and is of relevance to trading strategies based on assumptions of underlying $\alpha$-stable random behavior. In this direction, fractional Black-Scholes equations have been introduced (\cite{CARTEA2007749}, \cite{Kleinert_PhysicaA}) as appropriate models for hedging, since standard Black-Scholes theory is based on assumptions of normality/log-normality of the underlying processes. The example here can serve as a building block to applying fractional Black-Scholes theory to financial data.

\section{Conclusion}
\label{conclusion}
The Gaussian processes methodology based on the implementation of fractional derivatives and stationary covariance kernels in Theorem 
\ref{kernel_theorem} allows for the effective and efficient discovery of linear space-fractional equations in $\mathbb{R}^d$. We have demonstrated the feasibility, robustness, and accuracy of this methodology. Due to the versatilty of fractional archetypes and user-controlled parsimony, we demonstrated the appeal of the method even for discovering integer-order equations, and discussed a novel approach to interpolating multi-term linear PDEs.
The methodology allows for versatile discovery of linear space-fractional differential equations in $\mathbb{R}^d$ in many applications where fractional or anomalous behavior is expected, as illustrated by the discovery of a fractional diffusion equation for relative stock performance. This can be used to calibrate $\alpha$-stable parameters from time series, and has potential impact in risk management and fractional Black-Scholes theory. 

{This work is an extension of \cite{RAISSI2017683} and \cite{2018JCoPh.357..125R} to the case of fractional order-operators and a broader class of covariance kernels, including the Mat\'ern class. Compared to more standard approaches for inverse and parameter estimation problems, the Gaussian process regression approach has the advantage that repeated forward solution of the differential equation is not required. Rather, the equation parameters are discovered by a single Gaussian process regression through scattered observations, which involves entirely standard maximum likelihood estimation using L-BFGS, albeit with a covariance kernel that is constrained by the differential equation. As discussed in  \cite{RAISSI2017683} and \cite{2018JCoPh.357..125R}, this constraint regularizes the Gaussian process and allows physical laws to be discovered with comparatively few data points. On the other hand, extension to space-dependent rather than constant coefficient fields, for example, presents a challenge, as strong prior knowledge of the coefficient field would be required in order to parametrize it and apply this methodology.} 

Regarding extensions of this work, one potential drawback of the methodology is that Gaussian processes may not scale well to large datasets. In this regard, remedies proposed in \cite{gps_for_big_data, 2017arXiv170403144R} may be worth exploring. On the other hand, neural networks are intrinsically suited to larger data sets.  Moreover, unlike Gaussian processes, neural networks do not require a linear relationship of the form \eqref{linear_relationship}, and can be freely used to discover parametrized nonlinear differential equations. Neural networks were used to solve and discover (integer-order) PDEs in \cite{2017arXiv171110561R, 2017arXiv171110566R}. The methodology therein was further utilized in \cite{2018arXiv180101236R} and \cite{2018arXiv180106637R} to train neural networks to distill the actual dynamics of nonlinear dynamical systems and nonlinear partial differential equations, respectively. See also \cite{2017arXiv171009668L} for a different approach.  In general, however, extension of these methods to allow for fractional-order differential operators would {benefit from} \emph{fractional automatic differentiation} of neural networks, which remains a major conceptual and numerical challenge due to a lack of classical chain rule for fractional-order operators.

\appendix
\section{Proof of Theorem \ref{kernel_theorem}}\label{appendix}
Recall the shift property of the Fourier transform:
\begin{equation*}
\mathcal{F}\{f(\bm{x}-\bm{a})\}(\bm{\xi}) = e^{-i\bm{a}\cdot\bm{\xi}}\mathcal{F}\{f(\bm{x})\}(\bm{\xi})  
\end{equation*}
By the stationarity of $k$,
\begin{equation*}
\mathcal{F}_{\bm{x}}\{k(\bm{x},\bm{y})\} = 
\mathcal{F}_{\bm{x}}\{K(\bm{x}-\bm{y})\} =
e^{-i\bm{y}\cdot\bm{\xi}}
\hat{K}(\bm{\xi})
\end{equation*}
By the multiplier property of $\mathcal{L}$, in Fourier space,
$\mathcal{L}_{\bm{x}} k$ is given by multiplying $\mathcal{F}_{\bm{x}} \left\{ k \right\}$ by the symbol ${m(\bm{\xi})}$: 
\begin{equation*}
\mathcal{F}_{\bm{x}} \{\mathcal{L}_{\bm{x}} k\} =
e^{-i\bm{y}\cdot\bm{\xi}}
m(\bm{\xi})
\hat{K}(\bm{\xi})
\end{equation*}
Taking the inverse Fourier transform gives
\begin{align}
\mathcal{L}_{\bm{x}} k  &= \mathcal{F}_{\bm{x}}^{-1}
\left\{
{\mathcal{F}_{\bm{x}} \{\mathcal{L}_{\bm{x}} k \}} 
\right\} \nonumber \\
&=
\mathcal{F}_{\bm{x}}^{-1}
\left\{
e^{-i\bm{y}\cdot\bm{\xi}}
m({\bm{\xi}})
\hat{K}(\bm{\tilde{x}})
\right\} \nonumber \\
&=
\label{FT_shift}
\frac{1}{\sqrt{2\pi}}
\int_{\mathbb{R}^d}
e^{i\bm{x}\cdot\bm{\xi}}
e^{-i\bm{y}\cdot\bm{\xi}}
m(\bm{\xi})
\hat{K}(\bm{\xi})
d\bm{\xi}
\end{align}
This is the first of formulas \eqref{kernel_theorem_equations}.
As for $\mathcal{L}_{\bm{y}} k$, define the transpose operator by
${F^T(\bm{x}, \bm{y}) = F(\bm{y}, \bm{x})}$. Then 
$\mathcal{L}_{\bm{y}} F = \left[ \mathcal{L}_{\bm{x}} F^T \right]^T$ 
and symmetry of $k$ implies
\begin{equation*}
\mathcal{L}_{\bm{y}} k =
\left[ \mathcal{L}_{\bm{x}} k^T \right]^T =
\left[ \mathcal{L}_{\bm{x}} k \right]^T.
\end{equation*}
%\begin{equation*}
%\left[ \mathcal{L}_{\bm{y}} k (\bm{x},\bm{y}) \right] (\bm{x}_0, \bm{y}_0)
%= \left[ \mathcal{L}_{\bm{x}} k (\bm{y},\bm{x}) \right] (\bm{y}_0, \bm{x}_0)
%= \left[ \mathcal{L}_{\bm{x}} k (\bm{x},\bm{y}) \right] (\bm{y}_0, \bm{x}_0),
%\end{equation*}
In other words, $\mathcal{L}_{\bm{y}} k (\bm{x},\bm{y}) =\mathcal{L}_{\bm{x}} k (\bm{y},\bm{x})$, 
and does not require a separate computation (this is true for any covariance kernel). 
So far, only a single $d$-dimensional integation is required, which is the benefit
of knowing the Fourier transform $\hat{K}$ of ${K}$ analytically. Next, we see how
the stationary property gives the formula for 
$
\mathcal{L}_{\bm{y}} \mathcal{L}_{\bm{x}} k.
$
By the multiplier property of ${\mathcal{L}}$, we have
\begin{equation*}
\mathcal{L}_{\bm{y}} \mathcal{L}_{\bm{x}} k = 
\mathcal{F}_{\bm{y}}^{-1} \left\{ \mathcal{F}_{\bm{y}}^{}  \left\{ 
\mathcal{L}_{\bm{y}} \left[ \mathcal{L}_{\bm{x}} k (\bm{x},\bm{y}) \right]
\right\}(\bm{\xi'})
\right\}
=
\mathcal{F}_{\bm{y}}^{-1} \left\{ 
m(\bm{\xi'}) \mathcal{F}_{\bm{y}} \{\mathcal{L}_{\bm{x}} k (\bm{x},\bm{y})  \}
(\bm{\xi'})
\right\}.
\end{equation*}
Let us examine the inner term $\mathcal{F}_{\bm{y}} \{\mathcal{L}_{\bm{x}} k (\bm{x},\bm{y})  \} (\bm{\xi'})$. 
The Fourier transform ${\mathcal{F}_{\bm{y}}}$ 
passes through the integral over ${\bm{\xi}}$ in the final formula 
\eqref{FT_shift}  for ${\mathcal{L}_{\bm{x}} k (\bm{x},\bm{y}) }$. Inside that integral,
${\mathcal{F}_{\bm{y}}}$ only sees a constant term (independent of $\bm{y}$)
times the complex exponential $e^{-i\bm{y}\cdot\bm{\xi}}$. Thus, 
$\mathcal{F}_{\bm{y}} \{\mathcal{L}_{\bm{x}} k (\bm{x},\bm{y})  \} (\bm{\xi'})$ reduces to a $\delta$-function in
 $\bm{\xi'}$:
\begin{align*}
\mathcal{F}_{\bm{y}} \{\mathcal{L}_{\bm{x}} k (\bm{x},\bm{y})  \} (\bm{\xi'}) &=
\frac{1}{\sqrt{2\pi}}
\int_{\mathbb{R}^d}
 \left[\sqrt{2\pi} \delta(\bm{\xi} + \bm{\xi}')\right]
e^{i\bm{x}\cdot\bm{\xi}}
m(\bm{\xi})
\hat{K}(\bm{\xi})
d\bm{\xi} \\
&= 
\int_{\mathbb{R}^d}
 \delta(\bm{\xi} + \bm{\xi}')
e^{i\bm{x}\cdot\bm{\xi}}
m(\bm{\xi})
\hat{K}(\bm{\xi})
d\bm{\xi}
\end{align*}
Therefore, we obtain the second of formulas \eqref{kernel_theorem_equations}:
\begin{align*}
\mathcal{L}_{\bm{y}} \mathcal{L}_{\bm{x}} k
&= 
\mathcal{F}_{\bm{y}}^{-1} \left\{
m(\bm{\xi'})
\int_{\mathbb{R}^d}
\delta(\bm{\xi'} + \bm{\xi})
e^{i\bm{x}\cdot\bm{\xi}}
m(\bm{\xi})
\hat{K}(\bm{\xi})
d\bm{\xi} \right\}\\
&=
\frac{1}{\sqrt{2\pi}}
\int_{\mathbb{R}^d}
e^{i\bm{y}\cdot\bm{\xi'}}
\left\{
m(\bm{\xi'})
\int_{\mathbb{R}^d}
\delta(\bm{\xi'} + \bm{\xi})
e^{i\bm{x}\cdot\bm{\xi}}
m(\bm{\xi})
\hat{K}(\bm{\xi})
d\bm{\xi} \right\}  d\bm{\xi'}\\
&=
\frac{1}{\sqrt{2\pi}}
\int_{\mathbb{R}^d}
\int_{\mathbb{R}^d}
e^{i\bm{y}\cdot\bm{\xi'}}
\delta(\bm{\xi}' + \bm{\xi})
e^{i\bm{x}\cdot\bm{\xi}}
m(\bm{\xi})
m(\bm{\xi'})
\hat{K}(\bm{\xi})
d\bm{\xi'} 
d\bm{\xi}\\
&=
\frac{1}{\sqrt{2\pi}}
\int_{\mathbb{R}^d}
e^{-i\bm{y}\cdot\bm{\xi}}
e^{i\bm{x}\cdot\bm{\xi}}
m(\bm{\xi}) m(\bm{-\xi})
\hat{K}(\bm{\xi})
d\bm{\xi}.
\end{align*}

\bibliographystyle{siamplain}
\bibliography{ref_v3}
\end{document}